\documentclass{article}
\usepackage[T1]{fontenc}     
\usepackage{ragged2e}        
\usepackage[protrusion=true,expansion=false]{microtype}  
\usepackage{graphicx}
\usepackage{multirow}
\usepackage{caption}
\usepackage{float}
\usepackage{booktabs}
\usepackage{multicol}
\usepackage{tablefootnote}

\usepackage{footnote}
\makesavenoteenv{figure*}

\usepackage{textgreek} 
\usepackage{tabularx}        
\usepackage[table,xcdraw]{xcolor}
\usepackage{amsmath,amssymb,amsfonts}
\usepackage{mathtools} 
\usepackage{amsthm}
\usepackage{mathrsfs}
\usepackage{algorithm}
\usepackage{algpseudocode}
\usepackage{algorithmicx}

\usepackage{textcomp}
\usepackage{pifont}
\usepackage{setspace,lineno}
\usepackage{listings}
\usepackage{soul}            
\sethlcolor{yellow}          
\usepackage[title]{appendix}
\usepackage{manyfoot}
\usepackage{xcolor}
\newcommand{\x}[1]{\textcolor{black}{#1}}
\newcommand{\xx}[1]{\textcolor{black}{#1}}
\newcommand{\xxx}[1]{\textcolor{black}{#1}}

\definecolor{customgreen}{HTML}{00B050}
\definecolor{captionblue}{HTML}{0070C0}  
\newcommand{\cmark}{\textcolor{green}{\ding{51}}} 
\newcommand{\xmark}{\textcolor{red}{\ding{55}}} 
\DeclareCaptionLabelFormat{boldblue}{\textbf{\textcolor{captionblue}{#1 #2}}}
\usepackage{titlesec}
\titleformat{\section}
  {\normalfont\Large\bfseries\color{captionblue}}{\thesection}{1em}{}
\titleformat{\subsection}
  {\normalfont\large\bfseries\color{captionblue}}{\thesubsection}{1em}{}
\titleformat{\subsubsection}
  {\normalfont\normalsize\bfseries\color{captionblue}}{\thesubsubsection}{1em}{}
\usepackage{geometry}
\usepackage{amsmath} 
\allowdisplaybreaks[4]
\usepackage[protrusion=true,expansion=false]{microtype}
\usepackage{hyperref}
\hypersetup{
    colorlinks=true,
    linkcolor=blue,
    citecolor=blue,
    urlcolor=blue
}

\title{A Fine-Grained Attention and Geometric Correspondence Model for Musculoskeletal Risk Classification in Athletes Using Multimodal Visual and Skeletal Features}

\author{
	Md. Abdur Rahman\textsuperscript{1,4,a}, 
    Wasimul Karim\textsuperscript{1,4,a}, 
	Mohaimenul Azam Khan Raiaan\textsuperscript{2,a,*}, \\
	Tamanna Shermin\textsuperscript{1}, 
	Md Rafiqul Islam\textsuperscript{3}, 
	Mukhtar Hussain\textsuperscript{3},  
	Sami Azam\textsuperscript{3,*} \\
	\small
	\textsuperscript{1}Department of Computer Science and Engineering, United International University, Dhaka, 1212, Bangladesh \\
	\small
	\textsuperscript{2}Department of Data Science and Artificial Intelligence, Monash University, 3800, Melbourne, Australia \\
	\small
	\textsuperscript{3}Faculty of Science and Technology, Charles Darwin University, Darwin, 0909, Australia \\
	\small
	\textsuperscript{4}Applied Artificial Intelligence and Intelligent Systems (AAIINS) Laboratory, Dhaka, 1217, Bangladesh\\
	\small
	\textsuperscript{a}Equal Contributions\\
	\small
	\textsuperscript{*}Corresponding Author: mohaimenul.raiaan@monash.edu; sami.azam@cdu.edu.au
}

\date{} 
\begin{document}


\justifying
\maketitle
\begin{abstract}  
\noindent  Musculoskeletal disorders pose significant risks to athletes, and early risk assessment is essential for prevention. However, most existing methods are designed for controlled settings and fail to reliably assess risk in complex environments due to their reliance on a single type of data. This research introduces ViSK-GAT (Visual-Skeletal Geometric Attention Transformer), a novel multimodal deep learning framework that classifies musculoskeletal risk using both visual and skeletal coordinate-based features. A custom multimodal dataset (MusDis-Sports) was created by combining images and skeletal coordinates, with each sample labeled into eight risk categories based on the Rapid Entire Body Assessment (REBA) system. ViSK-GAT integrates two innovative modules: the Fine-Grained Attention Module (FGAM), which refines \xx{intra-modal features through self-attention before fusion}, and the Multimodal Geometric Correspondence Module (MGCM), which enhances cross-modal alignment between image features and coordinates. The model achieved robust performance, with all key metrics exceeding 93\%. Probability distribution error metrics also showed a low \x{Root Mean Squared Error (RMSE)} of 0.1205 and a \x{Mean Absolute Error (MAE)} of 0.0156. ViSK-GAT consistently outperformed state-of-the-art (SOTA) deep learning backbones and showed its potential to advance \x{artificial intelligence}-driven musculoskeletal risk assessment and enable timely interventions in sports.
\end{abstract}

\vspace{0.5em}
\vspace{1em}

\section{Introduction}
Musculoskeletal disorders (MSDs) are a primary concern in athletic performance. MSDs pose significant risks that can affect movement efficiency, endurance, and long-term health \cite{park2024fostering}. These disorders arise from repetitive strain, improper biomechanics, and sustained postural imbalances, leading to muscle fatigue, joint stress, and chronic injuries \cite{hout2025association}. Given these challenges, posture plays an important role in athletes' performance, endurance, and injury prevention. Proper posture improves balance, stability, and coordination, key factors in executing precise movements. In contrast, poor posture can cause muscle imbalances and an increased risk of musculoskeletal injuries such as strains, sprains, and joint stress \cite{porto2024effect}. Therefore, effective movement management is essential not only for athletes but for individuals of all ages and levels of activity, as it supports long-term health, safety, and overall well-being. Researchers have continuously sought methods to design safer training environments and develop automated movement recognition techniques \cite{wang2024explainable}. However, addressing these risks requires a proactive approach to posture monitoring and correction, where sports ergonomics becomes essential. Despite high levels of athletic conditioning, many athletes develop postural disturbances from repetitive training \cite{grabara2015comparison}. These eventually lead to musculoskeletal degeneration and chronic conditions \cite{bankosz2020habitual, gawel2021effect}.

Several biomechanical assessment methods have been introduced to assess posture and the risk of work-related musculoskeletal disorders (WMSD). These include the Rapid Entire Body Assessment (REBA) \cite{hignett2000rapid}, Rapid Upper Limb Assessment (RULA) \cite{mcatamney2004rapid}, Ovako Working posture Analysis System (OWAS) \cite{kivi1991analysis}, Occupational Repetitive Action (OCRA) \cite{colombini2016risk}, among others. These methods quantitatively analyze human postures to improve the reliability and precision of assessments. Although originally developed for occupational ergonomics, several studies support the extension of these methods (e.g., REBA) from industrial ergonomics to athletic contexts \cite{hulme2019applying}. These methods have also been effectively applied in sports settings. For example, REBA is applied across academic institutions \cite{kazemi2016evaluation} and sports domains \cite{gorce2024musculoskeletal} to evaluate postural risks associated with musculoskeletal disorders and to quantify injury risks.

Recently, posture assessment has increasingly become automated. \x{Early sensor-based approaches used wearable inertial measurement units to continuously monitor body movements and detect musculoskeletal risk in occupational settings \cite{zhao2021wearable}. More recently, integrated frameworks combining inertial measurement units (IMUs) and surface electromyography have demonstrated real-time injury risk assessment directly in field conditions with athletes \cite{alzahrani2026realtime}. However, such systems introduced discomfort and calibration issues that interfered with natural athletic performance \cite{lopez2016wearable, schall2022wearable}. Vision-based methods addressed this by offering non-invasive alternatives, where 2D joint angles extracted from standard cameras were used to compute ergonomic risk scores automatically \cite{li2020novel}. Building on this, OpenPose-based frameworks extended the approach to 3D skeletal data, which enables strong postural assessment under occlusion and non-frontal viewing conditions \cite{kim2021ergonomic}. These methods are further advanced through fully automated pipelines that apply rule-based REBA and RULA protocols to OpenPose-derived keypoints, without the need for wearable sensors or manual annotation \cite{benharkat2026automatic}.}

\x{Similar vision-based pipelines were further applied to construction worker postures using OWAS-based classification \cite{seo2021automated}, and more recently to REBA-based risk prediction across multiple image sources \cite{doong2025predicting}. To capture richer postural information, multimodal approaches emerged that combine complementary data types. Feature-level fusion of sensor signals with deep learning models achieved over 90\% accuracy in adverse conditions \cite{xiahou2023feature}. Cross-modal self-attention between infrared and pressure data further improved sitting posture recognition in office settings \cite{zhang2023multimodal}. Multimodal pose estimation methods have also been explored in industrial settings, combining 2D detection with point cloud geometry for robust grasping and manipulation tasks \cite{xiong2026multimodal}.}

Despite these advances, many models were designed for controlled settings, which limits their applicability to the dynamic real-world environment \cite{kee2021comparison, zhang2023multimodal}. These challenges highlight the need for a more adaptable system capable of providing precise real-time posture assessments in athletic environments.

Thus, to address these critical limitations, we propose ViSK-GAT (Visual-Skeletal Geometric Attention Transformer), a multimodal framework designed for robust and interpretable posture risk assessment in athletic environments. While cross-attention is an established mechanism for multimodal fusion, its standard application often treats each modality as a single undivided block. This is suboptimal for fine-grained postural analysis, where local image features (e.g., a bent elbow) must be precisely linked to specific skeletal keypoints. To address this, ViSK-GAT introduces a structured, two-stage attention pipeline. The Fine-Grained Attention Module (FGAM) first performs intra-modal refinement, enhancing discriminative local features within the image data before fusion. This prepared representation is then processed by the Multimodal Geometric Correspondence Module (MGCM), which performs a spatially-aware cross-modal alignment. Unlike generic cross-attention, MGCM is designed explicitly to bridge the domain gap between visual texture and geometric coordinate spaces, ensuring that the model's fusion is grounded in the biomechanical reality of the human body. 

\begin{figure}[!ht]
    \centering
    \includegraphics[scale=0.15]{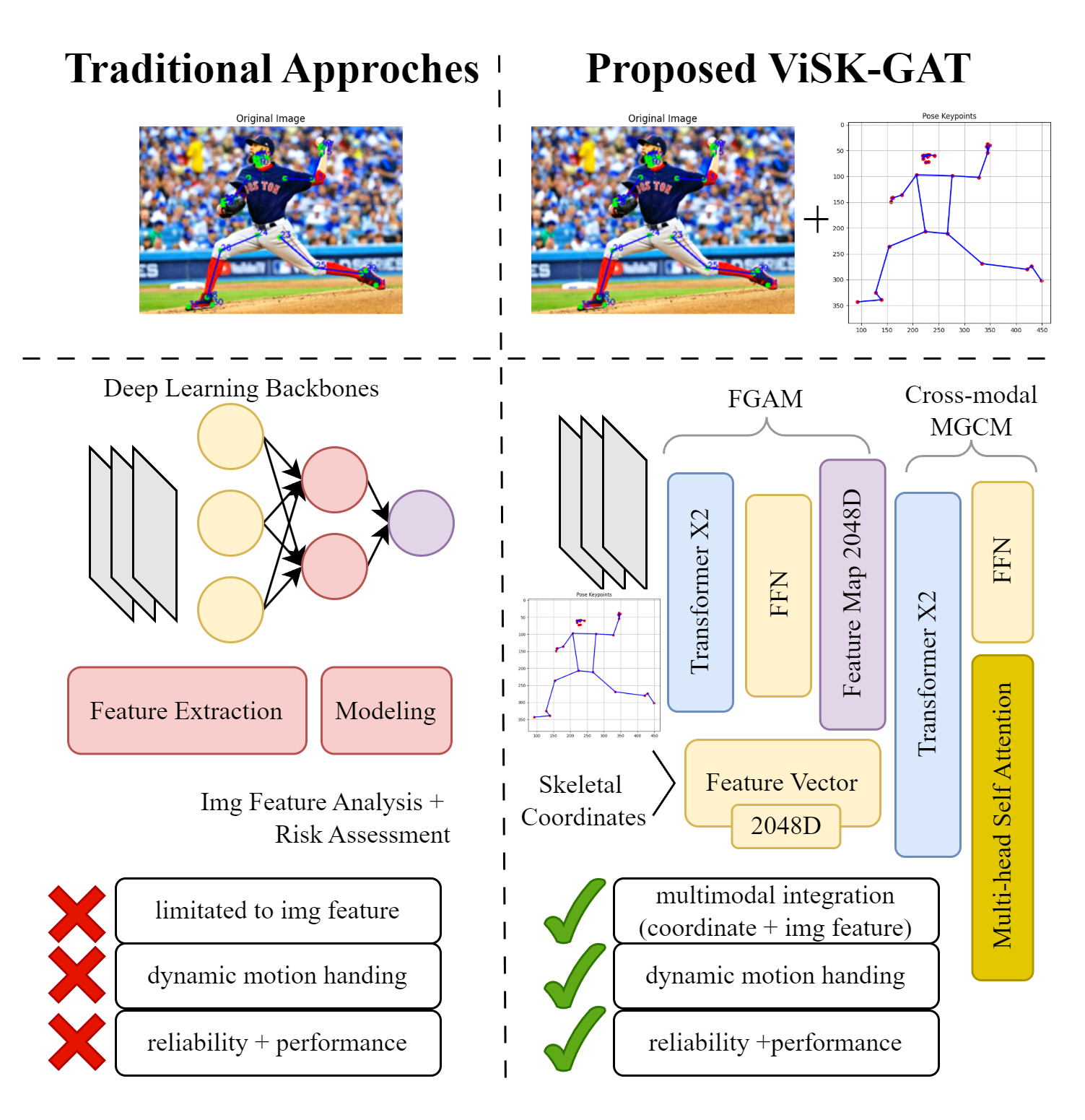}
    \caption{We processed the image features with Fine-Grained Attention Module (FGAM) and used a coordinate feature encoder for skeletal feature processing. The Multimodal Geometric Correspondence Module (MGCM) uses a cross-attention mechanism for the image and coordinate feature integration and risk assessment.}
    \label{fig:introDiagram}
\end{figure}

Furthermore, unlike conventional approaches (see Figure \ref{fig:introDiagram}) that struggle under real-world conditions, ViSK-GAT effectively combines high-level visual features with fine-grained skeletal data, enabling accurate posture classification even during dynamic movements and partial occlusion. Our extensive evaluations highlight the superiority of ViSK-GAT, which achieves a test accuracy of 93.89\%, significantly outperforming SOTA backbones, even when these models are paired with identical fusion and classification pipelines. Compared to previous studies on ergonomic risk assessment, ViSK-GAT demonstrates leading and competitive performance. Although prior models using ergonomic metrics such as OWAS and RULA report test accuracies ranging from 76.10\% to 93\%, ViSK-GAT surpasses these with a test accuracy of 93.89\%, also using the more fine-grained and anatomically detailed REBA scoring system, which is better suited for dynamic posture analysis in sports. These results collectively establish ViSK-GAT as a strong solution, both in terms of accuracy and architectural innovation, for multimodal ergonomic risk assessment, particularly in the complex and fast-paced conditions of athletic environments.

The major contributions of our work are as follows:
\begin{itemize}
\setlength{\itemsep}{0pt}
    \item A comprehensive multimodal annotated dataset is constructed by integrating multiple sports datasets and extracting human skeletal coordinates using MediaPipe. 
    \item A novel multimodal system is proposed that integrates visual and coordinate-based data to enhance the accuracy of the risk classification of musculoskeletal disorders in athletes, using REBA-based posture scoring as the core evaluation framework.
    \item A hybrid backbone is developed using Residual Blocks and Lightweight Transformer Blocks to extract features for efficient posture analysis in sports.
    \item A Fine-Grained Attention Module is introduced to refine image-based features through intra-modal self-attention before cross-modal fusion for more precise risk assessment.
    \item A Multimodal Geometric Correspondence Module is designed using a transformer-based framework to align images and coordinate features to ensure accurate representations of an athlete’s posture despite occlusions and perspective distortions.
\end{itemize}

The rest of the paper is organized as follows: Section \ref{RL} reviews related works on musculoskeletal disorder risk assessment with vision-based and multimodal approaches. Section \ref{methodology} describes the methodology, including the construction of the MusDis-Sports dataset and the design of the proposed multimodal framework. Section \ref{results} presents the experimental results, detailing the classification outcomes, configuration analysis, and comparison with recent studies. Section \ref{discussion} presents a discussion of the findings and their implications. Section \ref{lim_fwd} outlines the limitations and proposes directions for future research, and finally, Section \ref{conclusion} summarizes the main contributions and results of the study.

\section{Related works} \label{RL}
Numerous methods have been developed to help ergonomists assess the risks of Musculoskeletal Disorders (MSD). Traditional approaches, such as the Ovako Working posture Analysis System (OWAS), the Rapid Entire Body Assessment (REBA), the Musculoskeletal Disorders (MSD) Checklist, and the Rapid Upper Limb Assessment (RULA), have been widely adopted but often rely on manual scoring by ergonomic experts. Recent technological advances have enabled the integration of automated monitoring devices into MSD risk assessments.

\subsection{Vision-based methods}
Vision-based methods mostly use deep learning networks to analyze human activities with potential MSD risks in a non-invasive way. For instance, Li et al. \cite{li2020novel} proposed a vision-based real-time method to assess postural risk factors for musculoskeletal disorders using CNN feature extraction to generate 2D postures. The model achieved 93\% accuracy in detecting RULA action levels. \x{Kim et al. \cite{kim2021ergonomic} proposed an OpenPose-based system that extracts 3D skeletal data to compute joint angles and perform semi-automatic RULA/REBA postural assessments.} Seo et al. \cite{seo2021automated} introduced a computer vision-based method for the automatic identification of construction worker postures to assess ergonomic risk. Using classification techniques to learn from various postures depicted in virtual images, they achieved 89\% precision in classifying diverse postures in the images. 

\x{To improve flexibility in keypoint detection under constrained scenarios, Liu et al. \cite{liu2023ldcnet} introduced LDCNet for limb direction cues with a differentiated Cauchy encoding method to suppress positional uncertainty in 2D pose estimation. Extending skeleton-based coordinate encoding further, another study proposed EHPE \cite{liu2022ehpe}, an anisotropic Gaussian coordinate encoding model to describe skeletal direction cues between adjacent keypoints.} In another study, Dzeng et al. \cite{dzeng2017automated} proposed an approach that automatically tracks and categorizes postures based on OWAS using skeleton-based motion data to analyze ergonomic risks. The posture identification accuracies exceeded 90\% for most activities, with up to 88.5\% in the OWAS assessment. On the same assessment (i.e., OWAS), Yan et al. \cite{yan2017development} achieved a maximum accuracy of 89.2\% using three posture classifiers for construction hazard prevention. \x{Skeleton-based methods have also been applied to athlete movement classification; one such approach achieved over 96\% precision in real-time functional movement screening using pose-derived features \cite{wenbo2022skeleton}. More recently, a markerless pose estimation framework fused MediaPipe and Keypoint R-CNN to predict cyclist injury risk from joint angles without wearable sensors \cite{habeeba2026fusionpose}.} Many researchers have also proposed ergonomic assessment based on skeletons extracted either from 2D images from an ordinary camera or 3D images produced by RGB-D sensors \cite{yan2017development, doong2025predicting}. \x{Computer vision has also been applied to real-time workplace safety monitoring, where deep learning models detected unsafe operator behaviors in office environments \cite{barbosa2025dynamic}.}

\subsection{Multimodal methods}
While vision-based methods individually contribute valuable insights into MSD risk assessment, recent advances have considered the potential of combining multiple data modalities for posture recognition.

For instance, Xiahou et al. \cite{xiahou2023feature} proposed a feature-level fusion framework that integrates signals from various sensors using deep learning models such as multilayer perceptrons (MLPs), recurrent neural networks (RNNs), and long short-term memory (LSTM) networks. Their LSTM-based model maintained over 90\% accuracy even under adverse environmental conditions. Similarly, Zhao et al. \cite{zhao2022fast} designed a multimodal fusion algorithm that extracted motion features from both hand and leg movements using wavelet packet decomposition and sample entropy. Their method achieved recognition rates consistently above 95\%. Several studies have also combined visual and sensor data for posture and behavior recognition. For instance, Zhang et al. \cite{zhang2023multimodal} introduced a deep learning framework that fuses infrared and pressure map data to recognize sitting postures. Using modality-specific backbones and a cross-modal self-attention mechanism, they achieved an F1-score of 93.08\%. \x{Skeleton-based action recognition has also benefited from view-agnostic approaches, where adaptive 2D skeleton deformation was combined with a spatial-temporal graph convolutional network to handle viewpoint variations \cite{dong2022adaptive}.} 

Furthermore, Palash et al. \cite{palash2023emersk} introduced a modular two-stream CNN and RNN framework that fuses facial expressions, body posture, and gait features early, while utilizing situational knowledge to enhance multimodal emotion recognition. Meanwhile, Qi et al. \cite{qi2023fl} introduced a federated learning-based multimodal framework for fall detection, where time-series data from wearable sensors and visual data from cameras are fused at the input level using a Gramian Angular Field transformation. Their model achieved accuracies of 99.927\% and 89.769\% for binary and multi-class fall activity recognition tasks, respectively. 

\x{Multi-scale cross-feature aggregation has also been explored in transformer-based frameworks for gesture recognition. For instance, Narayan et al. \cite{narayan2025compact} developed a multiscale convolutional feature model that fused with a lightweight transformer to improve accuracy under occlusion and limited data conditions.} Similarly, Liu et al. \cite{liu2024multimodal} developed a hydrogel-based system for monitoring respiration and recognizing finger-pressing postures. Even though they achieved a posture recognition accuracy of 99.259\%, their work primarily relied on a wearable sensor system, which often adds additional inconvenience to the users. \x{In the sports domain, cross-modal action recognition has been addressed through self-adaptive uncertainty modeling, where robust unimodal features were sampled, and cross-modal interactions were refined to achieve 89.49\% recognition precision on egocentric sports datasets \cite{zhao2025sumrr}.}

Despite the notable advancements, existing studies still show drawbacks. For instance, vision-based methods, although non-invasive, often suffer from challenges such as limited camera coverage, occlusion, and sensitivity to environmental factors like lighting, which makes the posture recognition tasks less reliable in dynamic, fast-paced environments like sports. Although multimodal approaches show promise, many rely on manually designed fusion strategies and lack fine-grained cross-modal integration. Moreover, a significant portion of prior studies focus on specific activities or controlled environments, which reduces their applicability to diverse real-world settings. In this work, we propose a novel multimodal system that integrates both visual and coordinate-based data through fine-grained attention and geometric correspondence mechanisms and offers a more accurate and generalizable approach for musculoskeletal disorder (MSD) risk assessment in athletes.

\section{Methodology} \label{methodology}
This section details the dataset construction with coordinate extraction and REBA-based risk annotation, along with the design choices of our proposed model.

\subsection{MusDis-Sports dataset}
To support the training and evaluation of the proposed multimodal ergonomic risk assessment model, we constructed a domain-specific dataset (MusDis-Sports) for posture analysis in athletic scenarios. The image data were compiled from various publicly available sports datasets, including Sports-1M \cite{sabahesaraki2023sports1m}, Sports Classification \cite{gpiosenka2022sportsclassification}, Sports Image Classification \cite{sidharkal2022sportsimage}, and the Cricket Shot \cite{aneesh102022cricketshot} datasets. These sources were selected for their diversity in body configurations, motion states, and camera viewpoints to capture a wide range of postures.

\begin{figure*}[!ht]
\centering
\includegraphics[scale=0.65]{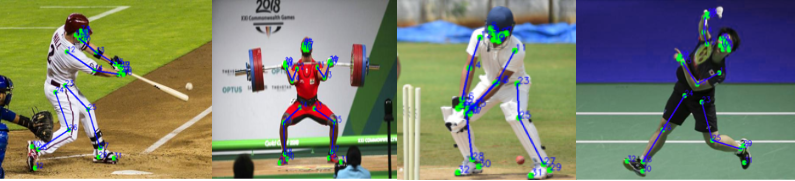} 
\caption{Sample images from the constructed MusDis-Sports dataset showing athletes in various sports-related postures with overlaid 2D skeletal keypoints extracted using the MediaPipe framework. These samples are used for training and evaluating musculoskeletal risk classification.}
\label{fig:dataset_overview}
\end{figure*}

Our MusDis-Sports dataset comprises 8989 annotated samples, each consisting of an image of a human subject in a sports-related posture, the corresponding 2D skeletal keypoints, and an ergonomically informed class label representing the REBA-based risk category. Figure \ref{fig:dataset_overview} visualizes some images from the constructed dataset. The dataset can be accessed through the footnoted GitHub repository\footnote{\url{https://github.com/mak-raiaan/MusDis-Sports_Dataset}}.

\subsubsection{Image and skeleton coordinate construction}
\label{img_coordinate_extn}
In this study, we curated a dataset (MusDis-Sports) of ergonomic postures in athletic contexts by selecting only those samples from publicly available sports datasets. In our study, only images with visible full-body configurations were retained. Each selected image was processed through the MediaPipe\footnote{\url{https://github.com/google-ai-edge/mediapipe}} framework to extract body landmarks. MediaPipe was selected as the pose estimation framework for this study due to its status as a widely adopted, open-source standard that provides a scalable and reproducible method for generating skeletal keypoints, which is necessary for large-scale dataset creation. This framework outputs 33 anatomical keypoints corresponding to major joints and body segments. For every subject present in an image, MediaPipe's pretrained model was used to extract 33 anatomical landmarks per subject, defined in Equation \eqref{eq:anatomical_landmark}:
\begin{equation}
\mathcal{K} = \{(x_i, y_i, v_i)\}_{i=1}^{33}, \quad (x_i, y_i) \in [0, 1]^2, \quad v_i \in [0, 1]
\label{eq:anatomical_landmark}
\end{equation}
where $(x_i, y_i)$ denote normalized 2D coordinates for landmark $i$, and $v_i$ represents the visibility confidence of that point. We filtered out unreliable landmarks by excluding any keypoint with $v_i < 0.5$. The remaining coordinates were scaled to the original image dimensions, forming the 2D skeletal matrix \({K} \in \mathbb{R}^{33 \times 2}\). To support downstream ergonomic analysis, angular features were extracted from skeletal keypoint triplets using two complementary methods.

First, we calculated the joint angles using the cosine law. For any three keypoints $A$, $B$, and $C$, the joint angle formed at $B$ was computed as shown in Equation \eqref{eq:cosine_angle}:
\begin{equation}
\theta_{ABC} = \cos^{-1} \left( \frac{(\vec{BA} \cdot \vec{BC})}{|\vec{BA}| \cdot |\vec{BC}|} \right)
\label{eq:cosine_angle}
\end{equation}
where $\vec{BA} = A - B$ and $\vec{BC} = C - B$ are vectors formed from the coordinates of the three points, and $\|\cdot\|$ denotes the Euclidean norm. With this, the angular displacement at joints such as the elbow, knee, or shoulder was computed.

Second, vertical inclination angles were computed to assess postural deviations like trunk or neck flexion. These were derived using a slope-based formulation that calculates the angular deviation from the vertical axis. For example, the trunk tilt angle $\theta_{\text{trunk}}$ between the shoulder and hip centers was given as shown in Equation \eqref{eq:trunk_inclination}:
\begin{equation}
\theta = \tan^{-1} \left( \xx{\frac{|x_{\text{shoulder}} - x_{\text{hip}}| + \epsilon}{|y_{\text{shoulder}} - y_{\text{hip}}|}} \right)
\label{eq:trunk_inclination}
\end{equation}
where $(x_{\text{shoulder}}, y_{\text{shoulder}})$ and $(x_{\text{hip}}, y_{\text{hip}})$ denote the 2D coordinates of the shoulder and hip midpoints, respectively. The small constant $\epsilon$ is included to prevent division by zero in near-vertical posture configurations. Algorithm \ref{alg:skeleton_coord} presents the overall process of extracting coordinates and calculating angles from a set of input images.

\begin{algorithm}[ht!]
\caption{Skeleton Coordinate Extraction and Angle Calculation Process}
\label{alg:skeleton_coord}
\begin{algorithmic}[1]
\State \textbf{Input:} images \(\mathcal{I} = \{I_1, I_2, \dots, I_N\}\)
\For{each image \(I \in \mathcal{I}\)}
    \State Apply MediaPipe to keypoints: \(\mathcal{K} = \{(x_i, y_i, v_i)\}_{i=1}^{33}\)
    \State Initialize empty keypoint list \(\mathbf{K}\)
    \For{each keypoint \((x_i, y_i, v_i) \in \mathcal{K}\)}
        \If{\(v_i \geq 0.5\)}
            \State Rescale \((x_i, y_i)\) to original image size
            \State Append to \(\mathbf{K}\)
        \EndIf
    \EndFor
    \For{each joint triplet \((A, B, C)\) in \(\mathbf{K}\)}
        \State vectors: \(\vec{BA} \gets A - B\), \(\vec{BC} \gets C - B\)
        \State angles: compute angle \(\theta_{ABC}\) 
    \EndFor
    \For{each body segment}
        \State v\_angle: vertical inclination angle \(\theta\)
    \EndFor
    \State Overlay keypoints and skeletal edges on image \(I\)
    \State Save the annotated image
\EndFor
\end{algorithmic}
\end{algorithm}

\subsubsection{REBA labeling}
We computed ergonomic risk scores using a structured implementation of the REBA protocol, following the methodology proposed by Hignett and McAtamney \cite{hignett2000rapid}. REBA evaluates musculoskeletal disorder risk from body posture and is widely adopted in ergonomics research \cite{jiao2024improved}. The process involves segmenting the human body into functional regions (e.g., trunk, neck, legs, upper arms, lower arms, and wrists) and assigning each part a discrete risk score based on the estimated joint angles.

In our work, once the angular displacements for relevant joints were computed (see Section \ref{img_coordinate_extn}), we assigned region-specific scores based on angle thresholds inspired by the REBA guidelines. Notably, the REBA scores are generated by applying a rule-based algorithm to the coordinates. The model therefore learns to predict the output of this function from visual and skeletal inputs, rather than learning an unconstrained mapping. We used a group-based formulation \cite{hignett2000rapid} in which the trunk, neck, and leg scores were aggregated to derive the Group A score, while the upper arm, lower arm, and wrist scores were combined to obtain the Group B score. These two group scores were subsequently mapped to a final REBA risk level through a predefined matrix, corresponding to the Group C scoring table outlined in the original REBA protocol. 

Our study specifically focuses on the six body parts: neck, trunk, wrist, upper arm, lower arm, and legs. As our dataset consists of single-frame images without dynamic tasks or load handling, we fixed both the load and activity modifiers to zero. The final Group C value was treated as the REBA scores, and based on the final calculated scores, we defined 8 classes, where classes with a higher number (i.e., 4$\leq$) indicate riskier postures. Table \ref{tab:dataset_size} reports the class-wise sample distribution across eight classes.

\begin{table}[!ht]
\centering
\caption{Class-wise distribution of annotated samples in the proposed MusDis-Sports dataset. Higher REBA \xxx{classes (e.g., Class 4–8)} indicate riskier postures associated with greater musculoskeletal disorder risk.}
\label{tab:dataset_size}
\begin{tabular}{ccc}
\midrule
\textbf{REBA Class} & \textbf{Samples} & \textbf{Ratio (approx.)}\\
\midrule
Class 1 & 1083 & 12.05\% \\
Class 2 & 1213 & 13.49\% \\
Class 3 & 1181 & 13.14\% \\
Class 4 & 1332 & 14.82\%\\
Class 5 & 1144 & 12.73\% \\
Class 6 & 1117 & 12.43\% \\
Class 7 & 888 &  09.88\% \\
Class 8 & 1031 & 11.47\% \\ 
\midrule
Total   & 8989 & 100\%\\
\midrule
\end{tabular}
\end{table}

Although some REBA protocols rely on 3D angular measurements, we employed a 2D approximation for broader applicability in conventional available image data that preserves the relative ranking of joint stresses. The resulting classes, therefore, represent postural intensity levels derived from 2D angular displacements. This reinterpretation ensures the labels remain meaningful even within monocular visual data. To ensure labeling reliability despite the automatic nature of REBA score generation: (i) \xx{we used MediaPipe’s built-in visibility confidence score ($v_i < 0.5$) to exclude low-certainty landmarks}; (ii) for each body-region score, we analyzed internal consistency by correlating the derived angular magnitudes across overlapping joints (e.g., shoulder–elbow–wrist chains) and retained samples with consistent geometry; and (iii) Furthermore, to assess labeling stability, we performed a manual sanity check on a random subset of 500 images to visually confirm that the overlay of skeletal keypoints was accurate and that the assigned REBA class was plausible given the posture. We also processed the samples under slightly varied viewing scales and compared them using the intraclass correlation coefficient (ICC = 0.84). Our findings indicate that the automatically derived REBA labels are self-consistent within a reasonable tolerance for 2D estimation. The REBA score assignment process is outlined in Algorithm \ref{alg:reba}.

\begin{algorithm}[ht!]
\caption{REBA Score Assignment from Posture Scores}
\label{alg:reba}
\begin{algorithmic}[1]
\State \textbf{Input:} Scores: $s_{\text{neck}}$, $s_{\text{trunk}}$, $s_{\text{leg}}$, $s_{\text{upper}}$, $s_{\text{lower}}$, $s_{\text{wrist}}$
\Function{GroupA}{$s_{\text{neck}}, s_{\text{trunk}}, s_{\text{leg}}$}
    \State use REBA lookup to compute $g_A$
    \State \Return $g_A$
\EndFunction
\Function{GroupB}{$s_{\text{upper}}, s_{\text{lower}}, s_{\text{wrist}}$}
    \State use REBA lookup to compute $g_B$
    \State \Return $g_B$
\EndFunction
\Function{GroupC}{$g_A, g_B$}
    \State $S_{\text{REBA}} \leftarrow \text{table}[g_A][g_B]$
    \State \Return $S_{\text{REBA}}$
\EndFunction
\State $g_A \leftarrow$ \Call{GroupA}{$s_{\text{neck}}, s_{\text{trunk}}, s_{\text{leg}}$}
\State $g_B \leftarrow$ \Call{GroupB}{$s_{\text{upper}}, s_{\text{lower}}, s_{\text{wrist}}$}
\State $S_{\text{REBA}} \leftarrow$ \Call{GroupC}{$g_A, g_B$}
\State \Return $S_{\text{REBA}}$
\State \textbf{Output:} final REBA score, $S_{\text{REBA}}$
\end{algorithmic}
\end{algorithm}

\subsection{Proposed model}
The proposed Visual-Skeletal Geometric Attention Transformer (ViSK-GAT) is a novel multimodal architecture that effectively integrates visual and coordinate-based skeletal representations using transformer-enhanced attention mechanisms for robust classification. 

The model begins with a custom ResNet-inspired encoder backbone composed of residual blocks to extract high-dimensional visual features from input images. These features are then refined through two core modules that represent the key innovations of our approach. The Fine-Grained Attention Module (detailed in Section \ref{FGAM}) employs multi-head self-attention combined with lightweight transformer blocks to perform intra-modal refinement \xx{of image features before fusion.} This module enhances the ability of the model to focus on discriminative visual patterns while maintaining spatial consistency through residual connections and layer normalization. 

\begin{figure*}[ht!]
\centering
\includegraphics[scale=0.11]{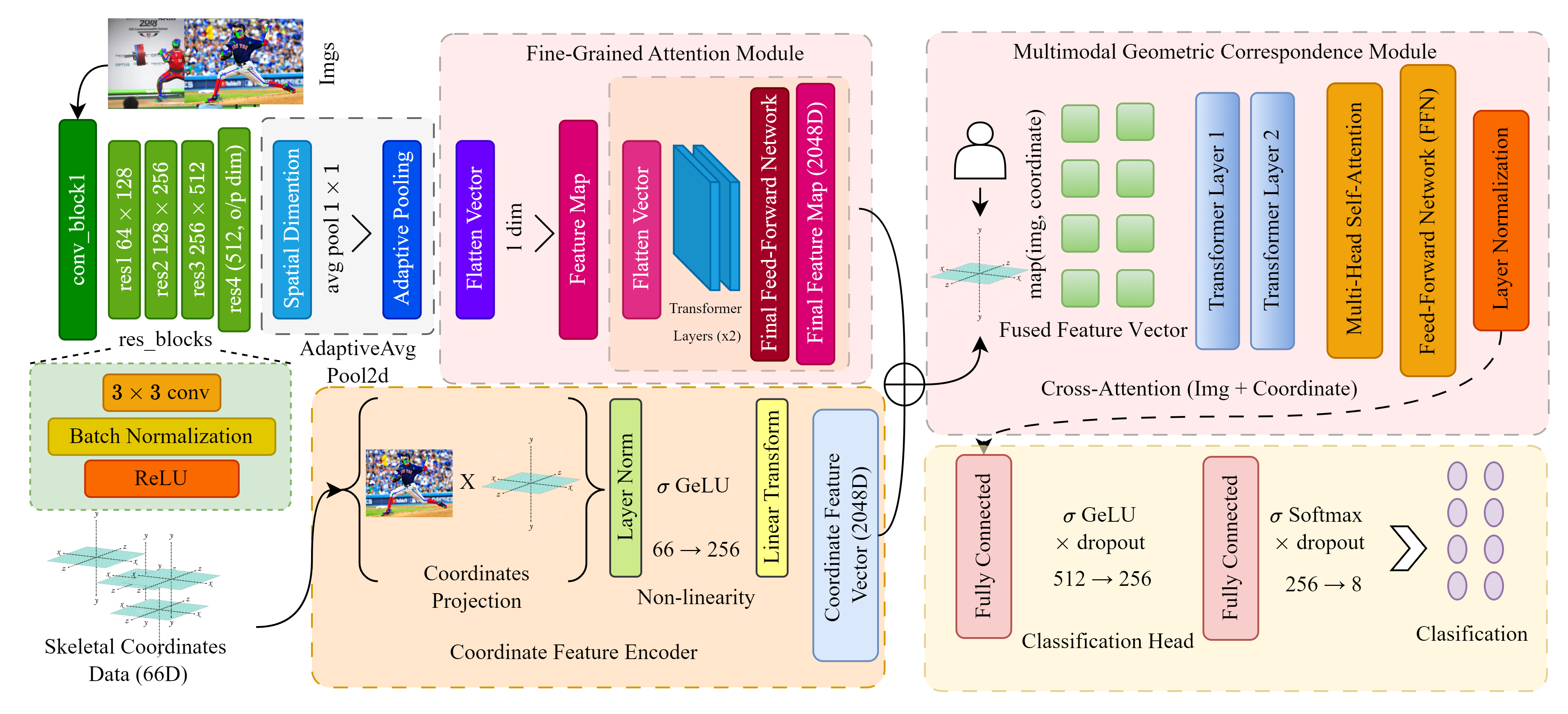} 
\caption{Overview of the proposed multimodal framework (ViSK-GAT) for ergonomic posture risk assessment. The system integrates visual features and skeletal coordinates through a hybrid backbone, enhanced by a Fine-Grained Attention Module and a Multimodal Geometric Correspondence Module.}
\label{fig:OverallModel}
\end{figure*}

The most significant contribution of the framework is the Multimodal Geometric Correspondence Module (MGCM), which aligns visual features with coordinate-based skeletal representations through cross-attention mechanisms. MGCM maps both modalities into a shared latent space, where attention-based feature alignment allows the model to learn complex cross-modal correspondences and geometric relationships. This alignment produces a coherent understanding of spatial structures and visual semantics. We further enhance the integrated features using a final transformer-based refinement stage, which captures long-range contextual dependencies across modalities. This leads to a unified and discriminative multimodal representation suitable for complex musculoskeletal risk classification tasks. Figure \ref{fig:OverallModel} illustrates the proposed ViSK-GAT framework.

\subsubsection{Hybrid multimodal feature extraction}
\label{HMFE}
To process both visual and coordinate data, we designed a dual-branch feature extraction backbone that handles the heterogeneity of input modalities. This module takes two distinct types of input: image frames representing the full body posture and the corresponding 2D skeletal keypoints.

First, the image branch operates directly on the image frames. We initially resized all frames to a fixed spatial resolution and passed them through Residual Blocks. We used the residual connections to maintain spatial information across deeper layers. Each residual block refines the image representation without suffering from vanishing gradients, ultimately generating a high-dimensional feature map $\xx{{F}_\text{img}} \in \mathbb{R}^{H \times W \times C}$, where $H$ and $W$ are the spatial dimensions, and $C$ is the channel depth, respectively.
Parallel to this, the coordinate branch handles structured skeleton data, where each frame contains 33 keypoints representing body joints. Each keypoint has 2D coordinates $(x_i, y_i)$, normalized to lie within $[0, 1]$. We began by flattening the skeleton into a sequence of joint tokens ${P} \in \mathbb{R}^{N \times 2}$, which was then projected to a higher-dimensional space using a learnable linear transformation, as shown in Equation \eqref{eq:e_pose}:
\begin{equation}
    \xx{{F}_\text{pose}^{\text{emb}}} = P {W}_e + {b}_e 
    \label{eq:e_pose}
\end{equation}
where ${W}_e \in \mathbb{R}^{2 \times d}$ and ${b}_e \in \mathbb{R}^{d}$ are the learned weights and biases for pose embedding, and $d$ is the latent dimension. The resulting pose embedding tensor $\xx{{F}_\text{pose}^{\text{emb}}} \in \mathbb{R}^{N \times d}$ represents spatial relationships among joints. Then, to capture higher-order dependencies and long-range interactions between postures, we passed \xx{${F}_\text{pose}^{\text{emb}}$} through a lightweight Transformer block, which uses self-attention to model correlations between distant joints and output an enhanced coordinate feature representation, \xx{${F}_\text{pose}$}. The Transformer block includes multi-head attention and feed-forward layers, allowing the model to learn contextual interactions between joints, such as whether the arms and torso are aligned in a squat or the symmetry of the legs during a lunge.

Finally, the output of both branches (i.e., \xx{${F}_\text{img}$} and \xx{${F}_\text{pose}$}) formed the basis for cross-modal reasoning in the following modules. We kept these representations distinct at this stage to maintain modality-specific semantics.

\subsubsection{Fine-grained attention module}
\label{FGAM}
After extracting modality-specific features from the dual branches (i.e., image tokens $\xx{{F}_\text{img}} \in \mathbb{R}^{256 \times 128}$, and pose embeddings $\xx{{F}_\text{pose}} \in \mathbb{R}^{33 \times 128}$), we found that the image representation alone was not sufficient to encode subtle postural variations relevant for risk classification. Although the backbone extracted general visual features effectively, it lacked contextual focus on regions important for posture interpretation, such as limb orientation or joint symmetry.

\begin{figure}[!ht]
\centering
\includegraphics[scale=0.13]{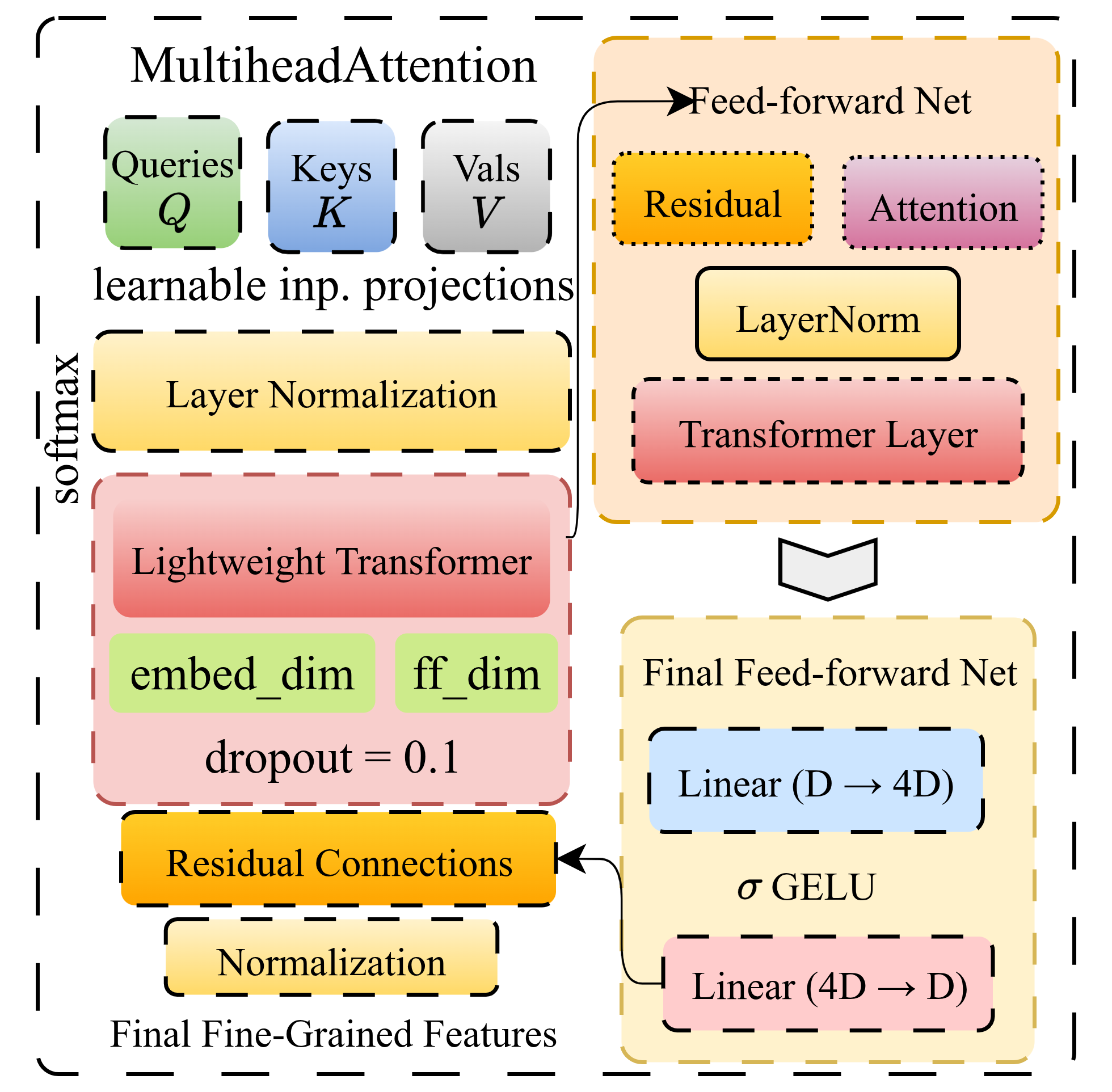} 
\caption{Architecture of the proposed Fine-Grained Attention Module for enhancing token representations.}
\label{fig:EnhancedFineGrainedAttentionModule}
\end{figure}

To address this, we introduced a Fine-Grained Attention Module (FGAM) for refining the token sequence (see Figure \ref{fig:EnhancedFineGrainedAttentionModule}). Our goal was to let each image patch attend to others within the same modality and enhance discriminative posture cues before engaging in cross-modal fusion. \xx{This intra-modal refinement step addresses a specific limitation of standard fusion approaches. Raw CNN features encode general visual patterns but lack sensitivity to postural cues such as limb orientation and joint asymmetry. Refining these features within the image modality before cross-modal fusion ensures that the alignment in MGCM operates on postural-cue-aware representations rather than on diffuse general-purpose features.}

We began by passing \xx{${F}_\text{img}$} through a multi-head self-attention layer with 8 attention heads. This initial attention layer allowed every token to attend to all others in the image sequence and modeled contextual dependencies across spatially distributed patches. To preserve the original representation, we added a residual connection followed by a layer normalization operation. This yielded an intermediate representation $F_1 \in \mathbb{R}^{256 \times 128}$.

However, this single-layer attention was seen as insufficient for modeling more complex long-range dependencies, such as posture-induced visual distortions. We stacked two Lightweight Transformer blocks to address this, where each was designed to refine the representation without incurring high computational cost. Each block then included multi-head self-attention, dropout, and a feed-forward sub-layer with expansion from 128 to 512 and back to 128 dimensions. The first block transformed $F_1$ into $F_2$, and the second further refined it into $F_3$. 

Finally, the refined tokens $F_3$ were passed through a position-wise feed-forward network to reweigh semantic features. We again used residual addition and normalization to ensure stability, producing the final enhanced image representation as shown in Equation \eqref{eq:f_img}:
\begin{equation}
    \hat{{F}}_\text{img} = \text{LayerNorm}\left( F_3 + \text{FFN}(F_3) \right)
    \label{eq:f_img}
\end{equation}
where \(FFN\) refers to the feed-forward network. This operation ensured that the learned image embeddings not only preserved spatial structure but also carried posture-specific attention cues.

\subsubsection{Multimodal geometric correspondence module}
\label{MGCM}
\begin{figure*}[ht!]
\centering
\includegraphics[scale=0.052]{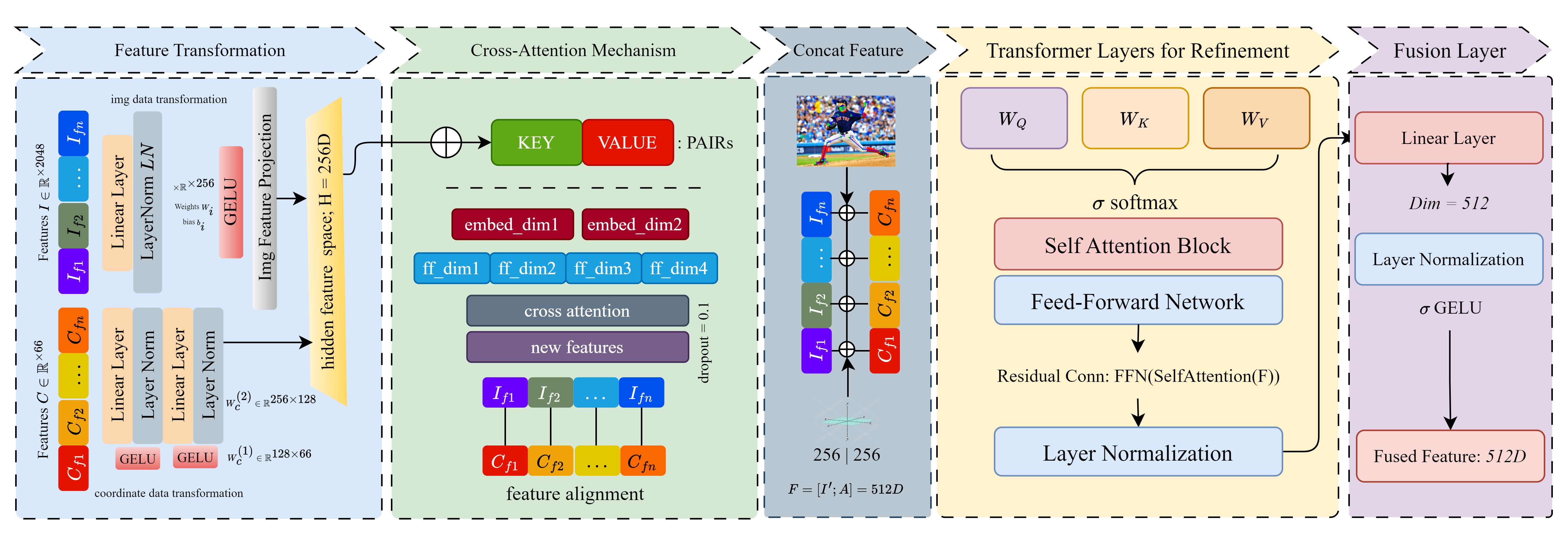} 
\caption{Proposed architecture of the Multimodal Geometric Correspondence Module (MGCM). This is a cross-attention transformer designed to bridge the gap between image tokens (Query) and skeletal coordinate embeddings (Key and Value). The process involves: (1) projecting both features into a common latent space; (2) using cross-attention to align visual appearance with structural posture information; (3) concatenating the attention output with the image feature; and (4) refining the combined feature using stacked Transformer layers to produce the final Fused Feature Vector ($F_{corr}$) for risk classification.}
\label{fig:EnhancedMultimodalGeometricCorrespondenceModule}
\end{figure*}

While image tokens encode rich texture and appearance cues, pose tokens capture structured joint relationships. To bridge this modality gap, we designed an enhanced Multimodal Geometric Correspondence Module (MGCM), which learns geometric correspondences using a cross-attention transformer architecture. \x{Visual texture and skeletal coordinates occupy fundamentally different representational spaces, and direct cross-attention between them without prior alignment can produce suboptimal correspondences. Image tokens serve as queries while coordinate tokens serve as keys and values, grounding the visual representation in the geometric structure of the body. This asymmetric formulation is motivated by the biomechanical nature of the task, where joint positions guide the selection of relevant visual features.} Figure \ref{fig:EnhancedMultimodalGeometricCorrespondenceModule} shows the overall architecture of the proposed geometric correspondence module.

We began by projecting both modalities into a common space. The image tokens were first passed through a linear transformation followed by LayerNorm and GELU activation to form ${F}^\prime_\text{img} \in \mathbb{R}^{256 \times d}$, where $d = 256$. Similarly, for the 33 anatomical keypoints, the coordinate embeddings were passed through a two-layer projection network with intermediate compression and re-expansion to produce ${F}^\prime_\text{pose} \in \mathbb{R}^{33 \times d}$. Following this, we applied a cross-attention mechanism, where image tokens acted as \textbf{Q}ueries, and coordinate tokens served as both \textbf{K}eys and \textbf{V}alues. This allowed each visual token to dynamically attend to relevant joint embeddings and enable the model to align appearance cues with structural posture information. The feature attention output of this interaction is defined in Equation \eqref{eq:f_att} as:
\begin{equation}
{F}_\text{attn} = \text{MultiHeadAttn}(Q={F}^\prime_\text{img}, K={F}^\prime_\text{pose}, V={F}^\prime_\text{pose})
\label{eq:f_att}
\end{equation}
where \({F}^\prime_\text{img} \in \mathbb{R}^{256 \times d}\) serves as the query matrix, and \({F}^\prime_\text{pose} \in \mathbb{R}^{33 \times d}\) is used as both the key and value matrices. We then concatenated ${F}^\prime_\text{img}$ with ${F}_\text{attn}$ to form a fused representation ${F}_\text{fused} \in \mathbb{R}^{1 \times 2d}$. To capture higher-order cross-modal interactions and enrich geometric understanding, we processed ${F}_\text{fused}$ through 2 stacked Transformer layers. Each layer consisted of multi-head self-attention with 4 heads, feed-forward blocks with dimensionality expanded to $4d$, and residual normalization. 

This transformer stack refined the fused representation and modeled long-range dependencies across modalities. Finally, the output token was passed through a fusion head comprising a linear layer, LayerNorm, and GELU activation to generate the joint embedding ${F}_\text{corr} \in \mathbb{R}^{512}$, used for final classification. This operation is summarized as shown in Equation \eqref{eq:f_corr}:
\begin{equation}
{F}_\text{corr} = \text{Fusion}\left\{ \text{Transformer}\left( [{F}^\prime_\text{img} \parallel {F}_\text{attn}] \right) \right\}
\label{eq:f_corr}
\end{equation}
The module ensured the model was sensitive to the geometric interplay between joint positions and visual appearance by aligning visual and structural cues through cross-attention and subsequent refinement. This representation was then passed to the classification head for posture risk prediction. \x{Figure \ref{fig:heatmap_viz} shows pose keypoints and their corresponding visibility confidence heatmaps. This visualization illustrates the spatial distribution of skeletal features that MGCM aligns with visual tokens through cross-attention. Joints with lower confidence, such as those near the extremities during dynamic movements, represent the cases where geometric alignment is most critical for accurate risk classification.}
\begin{figure*}[ht!]
\centering
\includegraphics[scale=0.1]{heatmap.png}
\caption{\x{Pose keypoints (top row) and skeletal keypoint confidence heatmaps (bottom row) for three representative samples from the dataset. Skeletal keypoints are detected with higher visibility confidence.}}
\label{fig:heatmap_viz}
\end{figure*}

\section{Results}
\label{results}
In this section, we present a comprehensive evaluation of the constructed MusDis-Sports dataset along with our proposed ViSK-GAT model for musculoskeletal risk assessment.

\subsection{Evaluation metrics}
To assess the performance of the proposed multimodal posture risk classification model, we employed a comprehensive set of evaluation metrics, including accuracy, precision, recall (sensitivity), and F1-score, as defined in Equations \eqref{eq:acc}-\eqref{eq:f1}:
\begin{equation}
\text{Accuracy} = \frac{TP + TN}{TP + TN + FP + FN}
\label{eq:acc}
\end{equation}
\begin{equation}
\text{Precision} = \frac{TP}{TP + FP}
\label{eq:prec}
\end{equation}
\begin{equation}
\text{Recall} = \frac{TP}{TP + FN}
\label{eq:recall}
\end{equation}
\begin{equation}
\text{F1-score} = 2 \cdot \frac{\text{Precision} \cdot \text{Recall}}{\text{Precision} + \text{Recall}}
\label{eq:f1}
\end{equation}
where TP, TN, FP, and FN represent true positive, true negative, false positive, and false negative, respectively. In addition to these metrics, we also calculated the Matthews Correlation Coefficient (MCC) and Cohen’s Kappa coefficient. MCC measures classification quality across all confusion matrix categories, with values ranging from -1 (total disagreement) to +1 (perfect prediction). The unweighted Cohen’s Kappa (\(\kappa\)) quantifies agreement between predicted and true labels, adjusted for chance, where \(\kappa\) = 1 indicates perfect agreement and \(\kappa\) = 0 indicates random agreement. The metrics are defined in Equations \eqref{eq:mcc}, and \eqref{eq:kappa}, respectively:
\begin{equation}
\small
\text{MCC} = \frac{TP \cdot TN - FP \cdot FN}{\sqrt{(TP + FP)(TP + FN)(TN + FP)(TN + FN)}}
\label{eq:mcc}
\end{equation}
\begin{equation}
\kappa = \frac{p_o - p_e}{1 - p_e}
\label{eq:kappa}
\end{equation}
where \(p_o\) is the observed agreement, and \(p_e\) is the expected agreement by chance. To quantify prediction deviation, we computed the Root Mean Squared Error (RMSE) and Mean Absolute Error (MAE) of the predicted probability distributions. These metrics measure the average deviation of the model's predicted probability vector from the one-hot encoded ground truth label, providing insight into the model's calibration and confidence in its predictions. The metrics are defined in Equations \eqref{eq:rmse} and \eqref{eq:mae}, respectively:
\begin{equation}
\text{RMSE} = \sqrt{\frac{1}{n}\sum_{i=1}^{n}||{\hat{p}}_i - {y}_i||^{2}_{2}}
\label{eq:rmse}
\end{equation}
\begin{equation}
\text{MAE} = \frac{1}{n}\sum_{i=1}^{n}||{\hat{p}}_i - {y}_i||_{1}
\label{eq:mae}
\end{equation}
where $n$ is the number of samples, ${\hat{p}}_i$ is the predicted probability vector (from the softmax output) for the $i^{th}$ sample, and ${y}_i$ is the one-hot encoded ground truth vector. In addition to these metrics, we further analyze the model's behavior across the posture risk levels using class-wise diagnostic metrics, namely Negative Predictive Value (NPV), False Positive Rate (FPR), False Discovery Rate (FDR), and False Negative Rate (FNR).

\subsection{Training analysis}
\label{training}
All experiments were conducted on an NVIDIA GeForce RTX 3060 GPU with 12GB of VRAM using CUDA 12.6. The system was powered by an AMD Ryzen 5 5600X processor with 12 threads clocked at 3.7 GHz. The software environment included Python 3.12.9, PyTorch 2.6.0+cu126, and cuDNN 90501.

The full multimodal dataset contained image-coordinate pairs annotated with REBA-based posture risk labels. \xx{The dataset was divided into 70\% training, 10\% validation, and 20\% testing sets. We trained our multimodal model for 100 epochs using a batch size of 16. The optimizer used was AdamW with a weight decay of $1 \times 10^{-5}$ to prevent overfitting. To improve learning dynamics, we used the OneCycleLR learning rate scheduler. This scheduler started with a low learning rate, gradually increased to a peak, and then decreased again over the course of training. The peak learning rate was set to $3\times 10^{-4}$, with a warm-up over 10\% of the epochs and a final decay using a decay factor of 1000. The model was trained on the resized (\(224\times224\)) image data using a categorical cross-entropy loss function.} At the end of training, the model achieved a training loss of 0.0007 and a training accuracy of 95.97\% in the final epoch. 

\subsection{Classification assessment}
\label{cls_asmnt}
To evaluate the performance of the proposed multimodal model, we assessed its ability to classify REBA scores based on posture. Our dataset includes a total of eight classes, each corresponding to a discrete REBA score ranging from 1 to 8. These labels were derived by analyzing the joint angles and positional configurations following the standard ergonomic assessment protocol \cite{hignett2000rapid}. 

\subsubsection{Class-wise performance} 
The model demonstrates consistently strong classification ability across all REBA classes, with F1-scores ranging from 88\% to 99\%. Table \ref{tab:class_metrics} summarizes these per-class performance metrics. 

\begin{table}[!ht]
\centering
\caption{Class-wise evaluation for REBA classification. \x{$\uparrow$ indicates better results. Best results are presented in bold.}}
\label{tab:class_metrics}
\small
\begin{tabular}{lcccc}
\midrule
\textbf{Class} & \textbf{Precision $\uparrow$} & \textbf{Recall $\uparrow$} & \textbf{F1-score $\uparrow$} & \textbf{Spec $\uparrow$} \\
\midrule
Class 1 & 95.85\% & 98.11\% & 96.97\% & 99.43\% \\
Class 2 & 88.46\% & 90.20\% & 89.32\% & 98.06\% \\
Class 3 & 91.25\% & 91.63\% & 91.44\% & 98.65\% \\
Class 4 & 89.47\% & 87.70\% & 88.58\% & 98.31\% \\
Class 5 & 95.49\% & 97.80\% & 96.63\% & 99.30\% \\
Class 6 & 96.07\% & 94.82\% & 95.44\% & 99.43\% \\
Class 7 & 98.79\% & 98.78\% & 98.78\% & \textbf{99.88}\% \\
Class 8 & \textbf{98.98}\% & \textbf{99.49}\% & \textbf{99.23}\% & \textbf{99.88}\% \\
\midrule
\end{tabular}
\end{table}

Notably, classes such as class 1 (low risk) and classes 7–8 (high risk) show the highest performance, with F1-scores of 96.97\%, 98.78\%, and 99.23\%, respectively. The specificity for these classes exceeds 0.99, indicating the model is very effective at ruling out incorrect classifications for extreme-risk postures. 

In contrast, classes 2 and 4 exhibit comparatively lower F1-scores of 89.32\% and 88.58\%, respectively, as these moderate-risk classes contain more ambiguous or transitional postures, which increase confusion with adjacent categories. We note that precision for class 2 is the lowest overall (88.46\%), and recall for class 4 is the lowest (87.70\%), suggesting these classes are the most challenging for the model. We also analyzed the multiclass ROC curves to further investigate the model’s discriminative ability across the eight REBA classes. 

\begin{figure*}[ht!]
\centering
\includegraphics[scale=0.35]{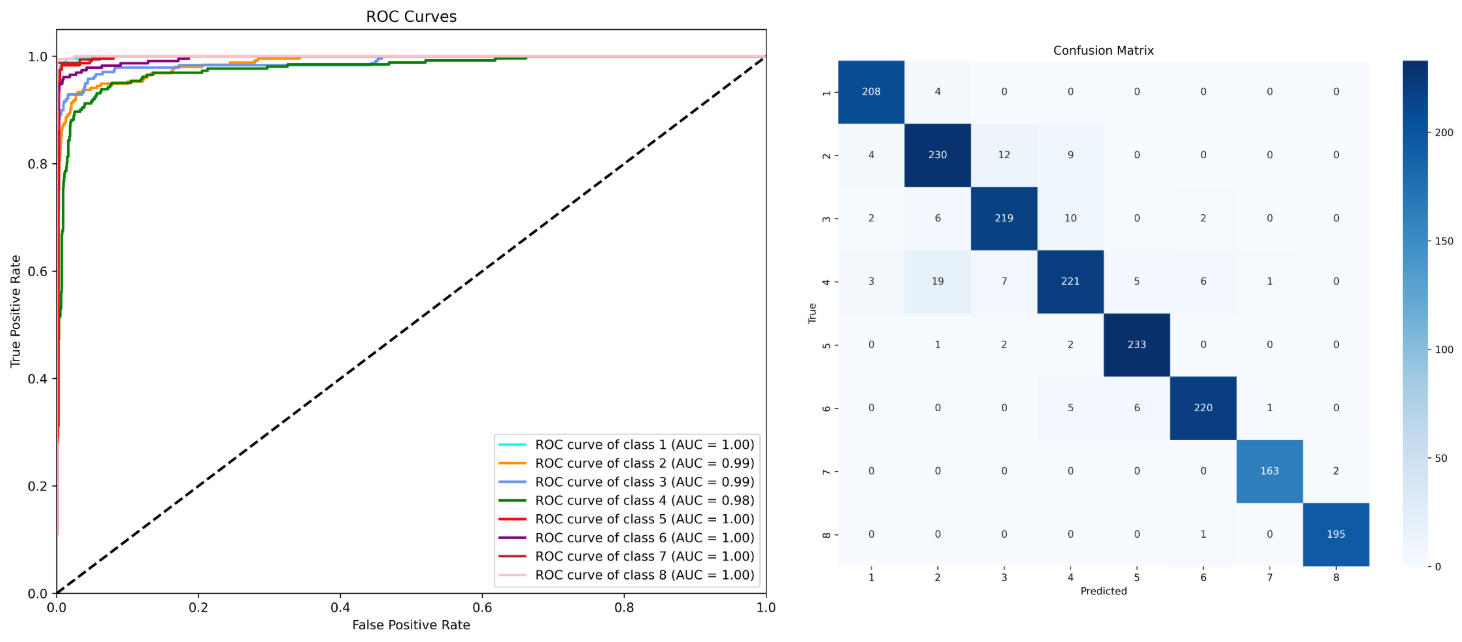} 
\caption{Left: Multiclass ROC curves for the eight REBA classes. Most classes achieve an AUC of 1.0, indicating excellent separability. Classes 2 and 3 scored 0.99, while class 4 scored 0.98. Right: Normalized confusion matrix showing classification performance across REBA classes. The strong diagonal pattern indicates high accuracy and minimal misclassification.}
\label{fig:roConf}
\end{figure*}

As shown in Figure \ref{fig:roConf} (left), the model shows excellent class separability, with most classes achieving an AUC of 1.0. Classes 2 and 3 scored slightly lower AUCs of 0.99, while class 4 recorded an AUC of 0.98. This observation is further supported by the confusion matrix in Figure \ref{fig:roConf} (right). 


\subsubsection{Overall evaluation} 
The model demonstrates superior musculoskeletal risk classification with an accuracy approaching 94\%. However, because accuracy alone can be misleading in multiclass settings, especially if class distribution is imbalanced, we further consider Cohen’s Kappa and the Matthews Correlation Coefficient (MCC), both of which are 0.93. These metrics adjust for chance agreement and class imbalance, respectively. The low RMSE and MAE values indicate that the model's predicted probability distributions are highly confident and closely aligned with the true labels, even for misclassified samples. The overall predicted probability distribution and result metrics are presented in Table \ref{tab:overall_metrics}.

\begin{table}[ht!]
\centering
\caption{Overall classification metrics and the predicted error probability distribution of the proposed model. \x{In the \textit{Metric} column, $\uparrow$ denotes higher is better; $\downarrow$ denotes lower is better.}}
\label{tab:overall_metrics}
\begin{tabular}{lc}
\midrule
\textbf{Metric} & \textbf{Value} \\
\midrule
Accuracy $\uparrow$ & 93.89\% \\
Cohen's Kappa $\uparrow$ & 0.93 \\
Matthews Correlation Coefficient $\uparrow$ & 0.93 \\
RMSE $\downarrow$ & 0.1205 \\
MAE $\downarrow$ & 0.0156 \\
\midrule
\end{tabular}
\end{table}

\subsubsection{Secondary error analysis} 
To identify where misclassifications are concentrated in the experiment, we examined Negative Predictive Value (NPV), False Positive Rate (FPR), False Discovery Rate (FDR), and False Negative Rate (FNR) for each class.

\begin{table}[ht!]
\centering
\caption{Class-wise values for Negative Predictive Value (NPV), False Positive Rate (FPR), False Discovery Rate (FDR), and False Negative Rate (FNR) for REBA classification. \x{$\uparrow$ denotes higher is better, and $\downarrow$ denotes lower is better. \xx{Best results for each metric are bolded.}}}
\label{tab:classwise_stats}
\begin{tabular}{lcccc}
\midrule
\textbf{Class} & \textbf{NPV $\uparrow$} & \textbf{FPR $\downarrow$} & \textbf{FDR $\downarrow$} & \textbf{FNR $\downarrow$} \\
\midrule
Class 1 & 0.9975 & 0.0057 & 0.0415 & 0.0189 \\
Class 2 & 0.9838 & 0.0194 & 0.1154 & 0.0980 \\
Class 3 & 0.9872 & 0.0135 & 0.0875 & 0.0837 \\
Class 4 & 0.9736 & 0.0169 & 0.1053 & 0.1565 \\
Class 5 & 0.9968 & 0.0070 & 0.0451 & 0.0210 \\
Class 6 & 0.9924 & 0.0057 & 0.0393 & 0.0517 \\
Class 7 & 0.9988 & \textbf{0.0012} & 0.0121 & 0.0121 \\
Class 8 & \textbf{0.9994} & \textbf{0.0012} & \textbf{0.0102} & \textbf{0.0051} \\
\rowcolor{gray!20}\textit{Avg.} & 0.9912 & 0.0088 & 0.0570 & 0.0558\\
\midrule
\end{tabular}
\end{table}

\begin{figure}[!ht]
\centering
\includegraphics[scale=0.25]{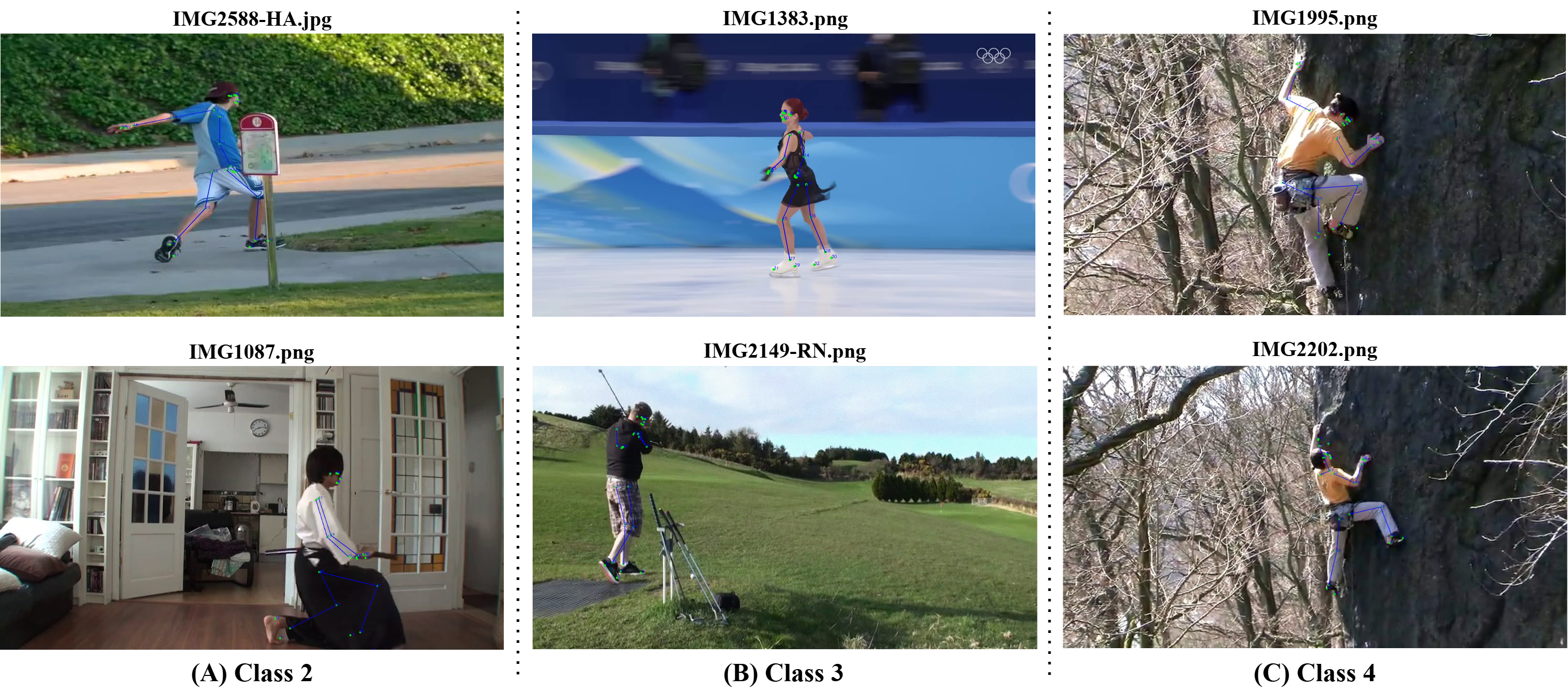} 
\caption{\xx{Representative failure cases from Classes 2, 3, and 4, where 2D skeletal projection leads to unreliable joint angle estimation. (A) Class 2: The subject's left side is not visible to the camera in both samples. In the second sample, loose-fitting garments further obscure the lower body joint configuration. Both conditions result in incomplete skeletal representations that fall near class boundaries. (B) Class 3: The first sample has motion blur from high-speed movement, where the subject's right side is absent from the frame. In the second sample, the subject faces away from the camera during a rotational posture, and trunk angle information is lost in the 2D projection.} (C) Class 4: The climber is pressed against a near-vertical rock surface where the torso and limbs overlap in the 2D projection. The camera angle is roughly perpendicular to the body, which further compresses depth information and distorts the estimated joint positions.}
\label{fig:failure_classes}
\end{figure}

As seen in Table \ref{tab:classwise_stats}, classes 7 and 8 show exceptional values, with FNRs of 0.0121 and 0.0051, and FDRs below 0.011, which means that these high-risk cases are both rarely missed and rarely falsely predicted. \xx{In contrast, class 2 struggles primarily with false alarms and records the lowest precision (88.46\%) and specificity (98.06\%), along with the highest FDR (0.1154) and second-highest FPR (0.0194). This class occupies a low-to-moderate risk region of the REBA score range where the angular thresholds separating it from Classes 1 and 3 are narrow. Small estimation errors in any single joint are therefore sufficient to shift the predicted label to an adjacent class, which can drive false predictions in both directions. Cases where this boundary proximity leads to misclassification are shown in Figure \ref{fig:failure_classes}(A).}

\xx{Class 3, while performing better than class 2, consistently sits in the lower half of the rankings across most error metrics, with an F1-score of 91.44\% and elevated FNR and FDR values of 0.0837 and 0.0875, respectively. This class also occupies a transitional region of the REBA score range where angular separation from adjacent classes is limited. Errors in trunk and upper limb angle estimation are consequently more impactful here than in higher-risk classes, where larger angular separations provide the model with greater tolerance for estimation noise. Figure \ref{fig:failure_classes}(B) provides illustrative examples.}

\xx{Among the eight classes, class 4 scored the weakest overall performance, with the lowest F1-score (88.58\%) and recall (87.70\%) in Table \ref{tab:class_metrics}, and the lowest NPV (0.9736) and highest FNR (0.1565) in Table \ref{tab:classwise_stats}. Since we rely on 2D skeletal projection, postures involving severe self-occlusion and non-standard viewpoints distort the estimated joint angles. Image backgrounds further interfere with reliable keypoint detection in such scenes. This combination of viewpoint distortion, threshold proximity, and detection noise caused the elevated FNR in this class. Figure \ref{fig:failure_classes}(C) shows representative cases.}

\xx{Despite these challenging cases, the model maintains strong overall performance across all eight classes, with a mean NPV of 0.9912, FPR of 0.0088, FDR of 0.0570, and FNR of 0.0558. The high-risk classes in particular are classified with exceptional reliability, which is the most critical requirement in a musculoskeletal risk assessment system.}

\subsection{Controlled baseline comparison}
\xx{For further experiment, we retrained existing representative methods directly on the MusDis-Sports dataset under identical experimental conditions for the 8-class REBA classification task (see Table \ref{tab:reproduced_comparison}).}

\xx{We experimented with various existing baseline architectures, including classical machine learning, hybrid skeleton pipelines, and advanced deep learning methods. For instance, we implemented the CNN-based pipeline\footnote{\url{https://github.com/LLDavid/RULA_2DImage}} of Li et al. \cite{li2020novel}, where 17 OpenPose keypoints are extracted and passed through a 4-layer fully connected network with batch normalization and ReLU activations. This configuration scored 87.42\% accuracy and 86.91\% F1-score with 23.50M parameters. We also reproduced the CrFe-CCT\footnote{\url{https://github.com/satyantazi/CrFe-CCT}} architecture of Narayan et al. \cite{narayan2025compact}, which uses multi-scale convolutional feature aggregation followed by a Compact Convolutional Transformer with sequence pooling. This model served as the strongest single-model baseline with 90.15\% accuracy and 89.73\% F1-score at 17.28M parameters.}

\xx{For the remaining methods, we constructed multiple plausible configurations. The silhouette-based SVM pipeline of Seo et al. \cite{seo2021automated} was tested across three feature configurations. EXP 1 relied on background-subtracted shape and ellipse features and produced the lowest accuracy of 74.18\%, due to the sensitivity of background subtraction to cluttered sports backgrounds. EXP 2 replaced this with MediaPipe-masked radial histogram features and produced 80.33\% accuracy, as the features were more robust to background variation. EXP 3 combined both feature sets and produced the best result of 84.67\% by capturing complementary shape and texture information, with parameter counts ranging from 8.10M to 8.58M across the three configurations. The view-invariant skeleton approach of Yan et al. \cite{yan2017development} was evaluated under three classifier settings. EXP 1 used only joint angle ratios with bagged trees and produced 76.91\%, while EXP 2 added length ratios to the feature set and produced 82.54\% by providing additional structural context. EXP 3 replaced bagged trees with a random forest and produced 85.89\% through better ensemble diversity and generalization, with parameters ranging from 9.45M to 10.24M.}

\begin{table*}[ht!]
\centering
\small
\caption{\xx{Comparison of ViSK-GAT with representative \xxx{multimodal baselines} retrained on our dataset. All models were trained under identical conditions (70:10:20 split, AdamW, $3\times10^{-4}$ LR, 100 epochs, 8-class REBA assessment). $\uparrow$ denotes higher is better; $\downarrow$ denotes lower is better. The best and the second-best results are bolded and underlined, respectively.}}
\label{tab:reproduced_comparison}
\renewcommand{\arraystretch}{1.8}
\resizebox{\textwidth}{!}{
\begin{tabular}{clllccccc}
\midrule
\textbf{Ref.} & \textbf{Author} & \textbf{Model} & \textbf{Key Implementation Details} & \textbf{Acc $\uparrow$} & \textbf{MAE $\downarrow$} & \textbf{F1 $\uparrow$} & \textbf{Params $\downarrow$} \\
\midrule
\cite{li2020novel} & Li et al. & CNN + 4-layer FC & OpenPose CNN $\rightarrow$ 17 keypoints $\rightarrow$ 4 FC layers (BN+ReLU) & 87.42\% & 0.0241 & 86.91\% & 23.50M \\
\cite{narayan2025compact} & Narayan et al. & CrFe-CCT & Multi-scale CNN $\rightarrow$ Compact Conv. Transformer $\rightarrow$ seq pooling & \underline{90.15\%} & 0.0198 & \underline{89.73\%} & 17.28M \\
\midrule
\cite{seo2021automated} & Seo et al. & Silhouette+SVM & \textit{EXP 1}: Bg-subtraction + shape/ellipse features + SVM & 74.18\% & 0.0472 & 72.85\% & 8.10M \\
& & & \textit{EXP 2}: MP-mask + radial histogram + SVM & 80.33\% & 0.0351 & 79.12\% & 8.35M \\
& & & \textit{EXP 3}: MP-mask + shape + radial histogram + SVM & 84.67\% & 0.0312 & 83.94\% & 8.58M \\
\cite{yan2017development} & {Yan et al.} & View-Invariant+Trees & \textit{EXP 1}: MP keypoints $\rightarrow$ angle ratios $\rightarrow$ Bagged Trees & 76.91\% & 0.0421 & 75.43\% & 9.45M \\
& & & \textit{EXP 2}: Angle + length ratios $\rightarrow$ Bagged Trees & 82.54\% & 0.0334 & 81.27\% & 9.89M \\
& & & \textit{EXP 3}: Angle/length ratios $\rightarrow$ Random Forest & 85.89\% & 0.0295 & 85.11\% & 10.24M \\
\cite{habeeba2026fusionpose} & {Habeeba et al.} & FusionPose+XGB & \textit{EXP 1}: MP only $\rightarrow$ joint angles $\rightarrow$ XGBoost & 81.76\% & 0.0378 & 80.52\% & 11.51M \\
& & & \textit{EXP 2}: MP + $v_i$ confidence weighting $\rightarrow$ XGBoost & 86.22\% & 0.0301 & 85.48\% & 11.78M \\
& & & \textit{EXP 3}: Weighted angles $\rightarrow$ XGB+LGBM ensemble & 88.94\% & \textbf{0.0117} & 88.31\% & 12.21M \\
\cite{barbosa2025dynamic} & {Barbosa et al.} & YOLO+MoveNet & \textit{EXP 1}: YOLOv8-n + MP keypoints $\rightarrow$ rule-based REBA & 79.45\% & 0.0401 & 77.89\% & \textbf{3.20M} \\
& & & \textit{EXP 2}: YOLOv8 feats + MP $\rightarrow$ MLP classifier & 84.91\% & 0.0327 & 83.76\% & \underline{4.10M} \\
& & & \textit{EXP 3}: YOLOv8-n + MP $\rightarrow$ RF classifier & 87.63\% & 0.0271 & 86.95\% & 4.90M \\
\rowcolor{gray!20} \multicolumn{2}{c}{Ours} & ViSK-GAT & FGAM + MGCM & \textbf{93.89\%} & \underline{0.0156} & \textbf{93.85\%} & 5.80M \\
\midrule
\end{tabular}
}
\end{table*}

\xx{We further assessed the FusionPose pipeline of Habeeba et al. \cite{habeeba2026fusionpose} across three XGBoost-based configurations of increasing complexity. EXP 1 used raw MediaPipe joint angles and produced 81.76\%, while EXP 2 introduced visibility confidence weighting to reduce the influence of unreliable keypoints and produced 86.22\%. EXP 3 combined XGBoost with a LightGBM ensemble and produced 88.94\% accuracy with the lowest MAE of 0.0117 across all compared methods, which reflects the strength of gradient boosting ensembles on well-structured tabular skeleton features, with parameters ranging from 11.51M to 12.21M across configurations. The detection-based framework of Barbosa et al. \cite{barbosa2025dynamic} was evaluated under three classification heads. EXP 1 applied rule-based REBA scoring directly on MoveNet keypoints and produced 79.45\%, as rigid thresholds struggled to generalize across the diversity of sports postures. EXP 2 replaced the rule-based head with an MLP classifier and produced 84.91\% through data-driven decision boundaries. EXP 3 used a random forest instead and produced 87.63\% through stronger resistance to overfitting, with the three configurations requiring 3.20M, 4.10M, and 4.90M parameters, respectively.}

\xx{As reported in Table \ref{tab:reproduced_comparison}, ViSK-GAT outperforms all reproduced baselines across all primary metrics. It achieves 93.89\% accuracy and 93.85\% F1-score and exceeds the best baseline while requiring only 5.8M parameters.}

\subsection{Comparison with state-of-the-art models}
To evaluate the effectiveness of the proposed multimodal framework, we conducted a comprehensive comparison against standard deep learning models widely used as image feature extractors. 

\xx{Given the nature of our dataset, we executed each of the CNN backbone models ourselves for image feature extraction}, while we processed the coordinate data using a multilayer perceptron (MLP). The MLP consisted of stacked fully connected layers with Rectified Linear Unit (ReLU) activation, layer normalization, and dropout. Features from both modalities were then fused through concatenation, followed by a shared classification head to predict the REBA posture risk score. We kept this configuration consistent across all models to isolate the effect of different image backbones. Table \ref{tab:backbone_comparison} summarizes the validation and test accuracy and loss values for each configuration.

\begin{table*}[ht!]
\centering
\small
\caption{Comparison of the proposed multimodal model with various standard image feature extractor backbones. $\uparrow$ denotes higher is better; $\downarrow$ denotes lower is better. \xx{The best and the second-best performance for each metric are bolded and underlined, respectively. All results were executed under identical training conditions on the proposed dataset. Parameter count is presented in millions (M)}.}
\label{tab:backbone_comparison}
\renewcommand{\arraystretch}{1.2}
\begin{tabular}{lccccc}
\midrule
\textbf{Backbone} & \textbf{Val Accuracy $\uparrow$} & \textbf{Val Loss $\downarrow$} & \textbf{Test Accuracy $\uparrow$} & \textbf{Test Loss $\downarrow$} & \textbf{Params (M) $\downarrow$} \\
\midrule
VGG16            & 83.04\% & 0.6985 & 83.13\% & 0.7273 & 14.71 \\
VGG19            & 84.12\% & 0.6844 & 84.20\% & 0.6911 & 20.02 \\
ResNet50         & 87.81\% & 0.5981 & 87.75\% & 0.6746 & 23.59 \\
ResNet101        & 88.44\% & 0.5832 & 88.39\% & 0.5812 & 42.66 \\
InceptionV3      & 87.03\% & 0.6114 & 87.02\% & 0.6105 & 21.80 \\
Xception         & 88.06\% & 0.5908 & 88.13\% & 0.5909 & 20.86 \\
EfficientNet-B0  & 87.45\% & 0.6043 & 87.42\% & 0.6129 & \underline{4.05} \\
MobileNetV2      & 84.25\% & 0.6732 & 84.23\% & 0.6714 & \textbf{2.26} \\
DenseNet121      & 88.12\% & 0.5925 & 88.06\% & 0.5947 & 7.04 \\
ViT-B/16         & 87.23\% & 0.5963 & 87.18\% & 0.6042 & 85.80 \\
Swin-T           & \underline{90.12}\% & \underline{0.5623} & \underline{90.08}\% & \underline{0.5714} & 27.52 \\
DeiT-S           & 88.71\% & 0.5814 & 88.65\% & 0.5893 & 21.67 \\
CvT-13           & 89.45\% & 0.5712 & 89.39\% & 0.5801 & 20.00 \\
BEiT-B           & 88.34\% & 0.5851 & 88.29\% & 0.5934 & 85.76 \\
\rowcolor{gray!20}ViSK-GAT (ours) & \textbf{93.66\%} & \textbf{0.4376} & \textbf{93.89\%} & \textbf{0.4808} & 5.80 \\
\midrule
\end{tabular} 
\end{table*}

Based on the results, CNN-based models show a clear performance progression with architectural depth, where deeper residual networks such as ResNet101 and DenseNet121 outperform earlier architectures such as VGG16 and MobileNetV2. Among the transformer-based backbones, Swin Transformer (Swin-T) achieves the highest baseline accuracy, followed by Convolutional Vision Transformer (CvT-13), Data-Efficient Image Transformer (DeiT-S), and BERT Pre-training of Image Transformers (BEiT-B). Vision Transformer (ViT-B/16) performs comparably to mid-range CNN models, as pure attention-based architectures tend to underperform on datasets of this scale without additional inductive biases. 

Despite these differences, all backbone combinations, including both CNN and transformer-based models, performed below our proposed model, since these backbones serve only as image encoders paired with simple concatenation and cannot replicate the intra-modal refinement of FGAM and the cross-modal alignment of MGCM. The lower training losses and higher validation scores of ViSK-GAT confirm more effective learning and generalization across all conditions. 

\xx{The parameter counts vary across models. Among the models, MobileNetV2 is the smallest at 2.26M, while ViT-B/16 and BEiT-B are the largest at around 85.8M. Most models fall in the 20M to 45M range. Our proposed model uses only 5.8M parameters, which is fewer than most models, yet still achieves the highest accuracy at 93.89\%.}

\x{To verify that the observed performance differences across all models are statistically significant, we applied a non-parametric Friedman test. The test ranks each model across multiple evaluation conditions and assesses whether the ranking pattern is consistent. For $k$ models evaluated across $n$ conditions, the Friedman statistic is computed as shown in Equation \eqref{eq:friedman}:
\begin{equation}
\chi^2_F = \frac{12n}{k(k+1)} \sum_{j=1}^{k} \left( R_j - \frac{k+1}{2} \right)^2
\label{eq:friedman}
\end{equation}
where $R_j$ is the mean rank of the $j^{th}$ model across all conditions. We evaluated all models across four evaluation conditions, as noted in Table \ref{tab:backbone_comparison}. The test achieved a $p$ value of $1.87 \times 10^{-5}$, which is well below the standard significance threshold of 0.05. This confirms that the performance differences among the models are statistically significant.}

\subsection{Ablation studies}
\subsubsection{Module configurations} 
Starting from a baseline model that used simple concatenation of image and coordinate features without any dedicated enhancement modules, we first introduced Residual Blocks (ResB) into the image encoder. This configuration resulted in an improvement in the F1-score of +0.0316 compared to the baseline (denoted as \(\Delta\)F1), along with an increase of +0.024 in Cohen's $\kappa$. This improvement is attributed to the deeper hierarchical representations enabled by residual connections, which help preserve low-level spatial details while allowing effective gradient flow. Building on that, we incorporated the Fine-Grained Attention Module (FGAM), which resulted in a $\Delta$F1 of +0.0501 and a \(\kappa\) gain of +0.044 compared to the baseline. FGAM allows the model to perform local, token-level attention within the image modality, rather than treating it globally. Building on FGAM, we added the Multimodal Geometric Correspondence Module (MGCM) to the framework. This led to the best results with a $\Delta$F1 of +0.0673 and \(\kappa\) improvement of +0.063 over the baseline. MGCM applies transformer-based cross-attention to explicitly model geometric relationships between modalities. \x{Table~\ref{tab:module_ablation} details the contribution of each key module and the intra-module design analysis.}

\begin{table*}[ht!]
\centering
\small
\renewcommand{\arraystretch}{1.2}
\caption{Ablation of key architectural modules. Performance improves incrementally with the addition of ResB, FGAM, and MGCM. \x{$\Delta$F1 in the module-level group is relative to M1 (initial baseline). $\Delta$F1 in the FGAM and MGCM groups is relative to M4 (final baseline). Abbreviations: SHCA = single-head cross-attention; MHCA = multi-head cross-attention; TransB = transformer block(s). Here, $\uparrow$ denotes higher is better, \xx{and the best results are \textbf{bolded}}.}}
\label{tab:module_ablation}
\begin{tabular}{lllcccccc}
\midrule
\textbf{Group} & \textbf{Exp.} & \textbf{Configuration} & \textbf{ResB} & \textbf{FGAM} & \textbf{MGCM} & \textbf{F1-score$\uparrow$} & \textbf{Cohen's $\kappa$$\uparrow$} & \textbf{$\Delta$F1$\uparrow$} \\
\midrule
{Module}
& M1 & Concat only (initial baseline)        & \xmark & \xmark & \xmark & 0.8712 & 0.867 & 0.00\% \\
& M2 & + ResBlocks                            & \cmark & \xmark & \xmark & 0.9028 & 0.891 & +3.16\% \\
& M3 & + FGAM                                 & \cmark & \cmark & \xmark & 0.9213 & 0.911 & +5.01\% \\
& M4 & + MGCM (final baseline)                & \cmark & \cmark & \cmark & \textbf{0.9385} & \textbf{0.930} & +\textbf{6.73}\% \\

{FGAM}
& A1 & \x{Initial self-attn only (no TransB)}  & \cmark & \cmark & \cmark & \x{0.9212} & \x{0.912} & \x{-1.73\%}\\
& A2 & \x{1 lightweight TransB}                & \cmark & \cmark & \cmark & \x{0.9298} & \x{0.921} & \x{-0.87\%}\\
& A3 & \x{2 TransB, 8 heads}                   & \cmark & \cmark & \cmark & \textbf{0.9385} & \textbf{0.930} & \textbf{0.00}\% \\
& A4 & \x{3 TransB, 8 heads}                   & \cmark & \cmark & \cmark & \x{0.9379} & \x{0.928} & \x{-0.06\%}\\

{MGCM}
& B1 & \x{No cross-attention (concat only)}    & \cmark & \cmark & \cmark & \x{0.9103} & \x{0.901} & \x{-2.82\%}\\
& B2 & \x{SHCA, w/o TransB}                    & \cmark & \cmark & \cmark & \x{0.9241} & \x{0.915} & \x{-1.44\%}\\
& B3 & \x{MHCA, w/o TransB}                    & \cmark & \cmark & \cmark & \x{0.9312} & \x{0.922} & \x{-0.73\%}\\
& B4 & \x{MHCA + 2 TransB}                     & \cmark & \cmark & \cmark & \textbf{0.9385} & \textbf{0.930} & \textbf{0.00}\% \\
& B5 & \x{MHCA + 3 TransB}                     & \cmark & \cmark & \cmark & \x{0.9371} & \x{0.927} & \x{-0.14\%}\\
\midrule
\end{tabular}
\end{table*}

\x{We have further extended the ablation to cover intra-module design choices within FGAM and MGCM, using M4 as the final baseline. In the FGAM design group, we notice that removing the transformer blocks entirely (A1) reduced F1 by -1.73\%. A single transformer block (A2) recovered most of this gap but still fell -0.87\% short, while adding a third block beyond the two transformer blocks offered no further gain.}

\x{On the other hand, in the MGCM fusion strategy group, replacing cross-attention with simple concatenation (B1) produced the largest drop of -2.82\%. This means that geometric alignment cannot be recovered through concatenation alone. Single-head cross-attention (SHCA, B2) and multi-head cross-attention without a transformer stack (MHCA, B3) closed this gap progressively at -1.44\% and -0.73\%, respectively. However, a third transformer layer (B5) showed marginal degradation consistent with overfitting on the compact 512-dimensional fused representation.}

\subsubsection{Hyperparameter configurations}
We conducted an extensive ablation study of 12 different experiments (Exp.) to evaluate the impact of various training hyperparameters on our multimodal posture classification model. The parameters tested include learning rate, batch size, weight decay, optimizer type, gradient clipping threshold, label smoothing, and loss functions. 

Experiments using adaptive optimizers such as Adam and AdamW generally outperformed SGD-based variants in both convergence and generalization. For example, Exp. 3 (SGD) showed fast per-epoch times, yet it struggled with higher losses. AdamW was more stable than vanilla Adam; as seen in Exp. 5 and Exp. 7, it contributed to smoother training dynamics and better generalization. The batch size of 16 proved to be an effective middle ground; configurations using larger batches (e.g., Exps. 1, 6, 9) trained faster but were slightly less robust, while smaller batches (Exps. 8, 12) led to very high training accuracy but also to signs of overfitting. Table \ref{tab:hyperparam_ablation} summarizes these configurations, along with corresponding training times per epoch, training and validation losses, and accuracies. 

\begin{table*}[ht!]
\centering
\renewcommand{\arraystretch}{1.2}
\caption{Ablation study on training hyperparameters: learning rate (LR), batch size, weight decay, optimizer, gradient clipping threshold (Clip), label smoothing (L/ Smooth), and loss function (Loss). Training inference time is reported in minutes per epoch (Inf. Time). The selected configuration for our final training is highlighted. \x{Here, $\uparrow$ denotes higher is better; $\downarrow$ denotes lower is better. \xx{The best and the second-best results are \textbf{bolded} and \underline{underlined}, respectively.}}}
\label{tab:hyperparam_ablation}
\footnotesize  
\begin{tabular}{c|ccclccl|ccc}
\midrule
\textbf{Exp} & \textbf{LR} & \textbf{Batch} & \textbf{Weight Decay} & \textbf{Optimizer} & \textbf{Clip} & \textbf{L/ Smooth} & \textbf{Loss} & \textbf{Inf. Time $\downarrow$} & \textbf{Loss $\downarrow$} & \textbf{Accuracy $\uparrow$} \\
\midrule
1  & 1e-3  & 32 & 1e-3   & Adam  & 1.0 & 0.1 & CrossEntropy   & $\sim$2:13 &        0.0018 & 0.9323 \\
2  & 1e-4  & 16 & 1e-4   & Adam  & 1.0 & 0.1 & Focal          & $\sim$1:45 &        0.0042 & 0.9080 \\
3  & 2e-4  & 32 & 1e-5   & SGD   & 0.5 & 0.1 & Focal & 		  $\sim$\textbf{0:50} &          0.0060 & 0.9162 \\
4  & 4e-4  & 16 & 1e-5   & Adam  & 1.0 & 0.2 & CE + Focal     & $\sim$1:57 &        0.0014 & 0.9365 \\
\rowcolor{gray!20}5  & 3e-4  & 16 & 1e-5   & AdamW & 1.0 & 0.1 & CrossEntropy   & $\sim$1:34 & 		\textbf{0.0007} & 0.9597 \\ 
6  & 3e-4  & 32 & 1e-4   & Adam  & 2.0 & 0.1 & CrossEntropy   & $\sim$\underline{0:51} &        0.0017 & 0.9284 \\
7  & 4e-4  & 16 & 1e-4   & AdamW & 1.0 & 0.1 & Focal          & $\sim$1:50 &        \underline{0.0010} & 0.9522 \\
8  & 3e-4  & 8  & 1e-5   & Adam  & 1.0 & 0.1 & CrossEntropy   & $\sim$2:40 &        0.0021 & \textbf{0.9995} \\
9  & 4e-4  & 32 & 1e-5   & AdamW & 1.0 & 0.1 & CE + Focal     & $\sim$1:10 &        0.0018 & \underline{0.9945} \\
10 & 3e-4  & 32 & 1e-5   & AdamW & 0.5 & 0.2 & CrossEntropy   & $\sim$0:58 &        0.0012 & 0.9391 \\
11 & 4e-4  & 16 & 1e-4   & Adam  & 1.0 & 0.1 & CrossEntropy   & $\sim$1:35 &        0.0015 & 0.9230 \\
12 & 3e-4  & 8  & 1e-5   & SGD   & 1.0 & 0.1 & CE + Focal     & $\sim$2:20 &        0.0050 & 0.9042 \\
\midrule
\end{tabular}
\end{table*}

Among the loss functions, standard CrossEntropy consistently performed well, whereas the combination with Focal loss (Exps. 4, 9, 12) slightly enhanced robustness in some cases but incurred longer training times. Using Focal loss alone (Exps. 2, 3, 7) helped with class imbalance to some extent, but was sensitive to the choice of optimizer. Moderate label smoothing (0.1) proved sufficient across most runs, while higher smoothing (0.2 in Exps. 4, 10) only offered marginal benefit. Similarly, clipping gradients above 1.0 (Exp. 6) led to reduced regularization and poorer generalization. While Exps. 8 and 9 achieved higher training accuracies of 0.9995 and 0.9945, respectively, their losses were relatively higher than that of Exp. 5. In contrast, Exp. 7 showed a competitive loss of 0.0010 but required longer training time and achieved slightly lower accuracy than Exp. 5. As a result, Exp. 5 was selected as the main configuration. It employed a learning rate of 3e-4, AdamW, a batch size of 16, and CrossEntropy loss with 0.1 label smoothing and gradient clipping of 1.0, and it provided the best overall performance.

\subsubsection{Features fusion}
To further validate our architecture, we compared ViSK-GAT against standard multimodal fusion baselines, holding the training data and hyperparameters constant. As shown in Table \ref{tab:ffusion}, simple fusion strategies like early and late fusion perform poorly, as they cannot model complex interactions. A standard cross-attention baseline performs better but is still outperformed by ViSK-GAT. This demonstrates that the specific, sequential refinement of FGAM followed by the geometric alignment of MGCM provides a significant advantage over a naive cross-modal approach.

\begin{table*}[ht!]
\centering
\renewcommand{\arraystretch}{1.1}
\small
\caption{Ablation of the fusion module. The results are compared with existing early fusion, late fusion, and the standard transformer encoder fusion strategies. \x{$\uparrow$ denotes higher is better.}}
\label{tab:ffusion}
\begin{tabular}{llcc}
\midrule
\textbf{Strategy} & \textbf{Fusion Method} & \textbf{Accuracy $\uparrow$} & \textbf{F1-Score $\uparrow$} \\
\midrule
Early Fusion & Pixel-coordinate concatenation at input & 71.31\% & 68.91\% \\
Late Fusion & Concatenation of final image/pose features & 84.23\% & 83.01\% \\
Cross-Attention & Standard transformer encoder fusion & 88.77\% & 86.51\% \\
\rowcolor{gray!20}ViSK-GAT (ours) & FGAM + MGCM & \textbf{93.89}\% & \textbf{93.85}\% \\
\midrule
\end{tabular}
\end{table*}

\subsubsection{Model complexity analysis}
We conducted a comprehensive analysis of our model's architectural footprint (see experimental hardware configuration in Section \ref{training}) to assess its suitability for real-world applications. The complete model comprises a total of 5.8 million parameters. The model requires 22.8 MB for trainable parameters and only 231 KB for non-trainable components, which results in a compact memory footprint suitable for various hardware platforms.

The model demonstrated efficient training processing a full epoch in 1:34 minutes (see inference details in Table \ref{tab:hyperparam_ablation}). The inference speed was measured at 8.2 milliseconds per frame, which translates to a processing rate of 122 frames per second. As detailed in Table \ref{tab:model_complexity}, the efficient training time, compact memory footprint, and rapid inference speed demonstrate that the model is computationally feasible for both development and deployment phases in real-time monitoring and assessment. 

\begin{table}[!ht]
\centering
\renewcommand{\arraystretch}{1.1}
\caption{Computational profile of the proposed ViSK-GAT model in terms of the major modules, including the Fine-Grained Attention Module (FGAM), and the Multimodal Geometric Correspondence Module (MGCM).}
\label{tab:model_complexity}
\small
\begin{tabular}{lcccc}
\midrule
\textbf{Component} & \textbf{Params} & \textbf{Trainable} & \textbf{Non-Trainable} \\
\midrule
Image Encoder            & 3.5 M & 14.2 MB & 121 KB \\
Coordinate Encoder       & 0.1 M & 0.4 MB  & 8   KB \\
FGAM                     & 1.2 M & 4.8 MB  & 63  KB \\
MGCM                     & 0.9 M & 3.2 MB  & 39  KB \\
Classification Head      & 0.1 M & 0.2 MB  &  0  KB \\
\rowcolor{gray!20}Total                    & 5.8 M  & 22.8 MB & 231 KB  \\
\midrule
\end{tabular}
\end{table}

\section{Discussion} 
\label{discussion}
In this work, our objective was to improve the classification of musculoskeletal risk in sports by addressing key limitations in current posture assessment methods: poor integration of visual and skeletal data, and limited adaptability to dynamic sports postures. We proposed ViSK-GAT, a multimodal deep learning framework featuring a hybrid architecture that integrates residual and transformer blocks, along with two novel modules: a Fine-Grained Attention Module (FGAM) \xx{for intra-modal image feature refinement} and a Multimodal Geometric Correspondence Module (MGCM) for better alignment between image and coordinate features. Furthermore, we introduced the MusDis-Sports dataset to support multimodal REBA-based classification in athletic settings.

We used the REBA scoring framework in our classification task to ensure ergonomic relevance and interpretability. To create a system capable of scalable deployment in diverse athletic environments, we based our analysis on 2D skeletal coordinates extracted using MediaPipe. This approach was strategically chosen because it allows the model to operate on standard monocular video, which is considerably more accessible and practical in real-world sports settings. From the 33 extracted 2D keypoints, we computed joint angles using cosine and slope-based methods. These angles were then mapped to region-specific REBA scores, which were combined through standard lookup tables to achieve the final risk score. While this 2D approximation introduces a known source of variation compared to a 3D biomechanical analysis, it provides a consistent and computationally efficient proxy for postural stress. Based on these scores, each sample was classified into one of eight discrete REBA risk classes, with higher class numbers indicating higher ergonomic risk.

To evaluate the contribution of each component to our architecture, we conducted stepwise ablation experiments, beginning with a baseline model that concatenated visual and skeletal features. As we incrementally incorporated residual blocks, the Fine-Grained Attention Module (FGAM), and the Multimodal Geometric Correspondence Module (MGCM), we observed consistent and significant improvements in performance. The complete ViSK-GAT model achieved a +6.73\% increase in the F1-score over the baseline, with a final test precision of 93.86\%, an F1-score of 93.85\%, and Cohen's Kappa of 0.93. These results reflect not only the effectiveness of our architecture but also its methodological novelty. Unlike existing approaches that often rely on simple fusion strategies or treat visual and skeletal modalities in isolation, ViSK-GAT introduces a structured multimodal integration pipeline that learns shared representations through attention-based mechanisms. MGCM enables explicit geometric alignment between image features and skeletal coordinates, allowing the model to understand spatial correspondences across modalities. \xx{This is complemented by FGAM, which performs intra-modal self-attention within the image modality to support fine-grained feature refinement.} Together, these modules contribute to a unified architecture capable of modeling complex, dynamic postures in sports environments. ViSK-GAT provides an effective approach for interpretable multimodal ergonomic analysis by combining transformer-based attention, geometric reasoning, and domain-specific REBA scoring.

\section{Limitations and future directions}
\label{lim_fwd}
Although our framework produced strong results, this study has a few limitations. We primarily focused on the critical task of classifying musculoskeletal risk from postural data (i.e., visual and skeletal coordinates), using a REBA-inspired scoring system based on joint angles and body segment positions. This approach provides a robust foundation for vision-based postural analysis by focusing on the most observable and measurable component of ergonomic risk.

The conventional protocol often incorporates additional modifiers for external load and repetitive activity. While these factors are important in a comprehensive risk assessment, they introduce variables that cannot be reliably estimated from visual data alone. Our model's design prioritizes high accuracy in the domain that computer vision can reliably address. Thus, the current model is more directly applicable to scenarios where risk is predominantly postural. \x{Additionally, the current framework relies on a CNN-based image encoder. While this choice was motivated by the need for real-time inference at a compact scale, integrating lightweight foundation models or LLM-based multimodal fusion strategies is a promising direction.} The most promising directions for future work include expanding the dataset to incorporate additional metadata such as sport type and athlete skill level, and \x{extending the framework to broader domains such as industrial ergonomics and daily activity monitoring, given the availability of comparable multimodal annotated datasets for those contexts.}

\section{Conclusion}
\label{conclusion}
In this study, we presented a multimodal deep learning framework (ViSK-GAT) to classify posture-related musculoskeletal risk in athletes by combining visual and skeletal features. Our key contributions include the design of a hybrid model architecture enhanced with a Fine-Grained Attention Module, which refines image-based cues. In addition, a Multimodal Geometric Correspondence Module improves alignment between image and coordinate data using cross-attention. We also introduced the annotated MusDis-Sports dataset, labeled with REBA scores derived from joint angle calculations based on MediaPipe skeletal keypoints. The proposed model achieved a test precision of 93.86\%, with an F1-score of 93.85\% and Cohen's Kappa of 0.93, and showed competitive and consistent performance at all levels of posture risk. Through detailed ablation studies, we showed that each architectural component contributed meaningfully to overall results, with the final configuration significantly outperforming the baseline models. In general, our framework provides an effective and interpretable solution for the classification of posture risk in sports environments. Its modular design and strong performance position it as a promising solution for broader ergonomic applications.

\section*{Declarations}
\noindent
\textbf{Conflict of Interests:} The authors declare that they have no financial conflicts of interest that could have influenced this work.\\
\textbf{Ethics Approval and Consent to Participate}. Not applicable.\\
\textbf{Data and Code Availability:} The dataset and code can be accessed through the following GitHub repository upon valid request for research purposes only: \url{https://github.com/mak-raiaan/MusDis-Sports_Dataset}.

\end{document}